\documentclass{article}
\pdfoutput=1

\usepackage{amsmath,amsfonts,bm}

\def\eqref#1{equation~\ref{#1}}
\def\1{\bm{1}}

\def\vmu{{\bm{\mu}}}
\def\vtheta{{\bm{\theta}}}
\def\vgamma{{\bm{\gamma}}}
\def\vbeta{{\bm{\beta}}}
\def\vphi{{\bm{\phi}}}
\def\vpsi{{\bm{\psi}}}
\def\vsigma{{\bm{\sigma}}}

\def\va{{\bm{a}}}

\def\vx{{\bm{x}}}
\def\vy{{\bm{y}}}

\DeclareMathAlphabet{\mathsfit}{\encodingdefault}{\sfdefault}{m}{sl}
\SetMathAlphabet{\mathsfit}{bold}{\encodingdefault}{\sfdefault}{bx}{n}

\def\sR{{\mathbb{R}}}

\newcount\colveccount
\newcommand*\colvec[1]{
        \global\colveccount#1
        \begin{bmatrix}
        \colvecnext
}
\def\colvecnext#1{
        #1
        \global\advance\colveccount-1
        \ifnum\colveccount>0
                \\
                \expandafter\colvecnext
        \else
                \end{bmatrix}
        \fi
}
\usepackage[T1]{fontenc}    % Use 8-bit T1 fonts

\usepackage{microtype}      % Microtypography
\usepackage[utf8]{inputenc} % Allow utf-8 input
\usepackage{hyperref}       % Hyperlinks
\usepackage{url}            % Simple URL typesetting
\usepackage{booktabs}       % Professional-quality tables
\usepackage{multirow}       % Helper package for organizing tables    
\usepackage{amsfonts}       % Blackboard math symbols
\usepackage{nicefrac}       % Compact symbols for 1/2, etc.
\usepackage{tabularx}       % Tables
\usepackage{makecell}       % More for tables
\usepackage{ragged2e}       % Justification
\usepackage{paralist}       % Compact lists
\usepackage{arydshln}       % Dashed lines in tables
\usepackage{xargs}          % Use more than one optional parameter in a new commands
\usepackage[colorinlistoftodos,prependcaption,textsize=tiny]{todonotes}
\usepackage{xparse}

\usepackage{import}         % Importing subdocuments
\usepackage{standalone}     % Compilable subdocuments and inclusion thereof
\usepackage{enumitem}
  \newlist{inlinelist}{enumerate*}{1}
  \setlist*[inlinelist,1]{%
          label=(\roman*),
      }

\usepackage{amsmath}
\usepackage{cases}          % Cases in math
\usepackage{amsthm}         % Theorems
\usepackage{thmtools}       % Tools for theorems
\usepackage{mathtools}      % More math commands
\usepackage{pgf}
\usepackage{tikz}
\usepackage{pgfplots}
	\usepgfplotslibrary{groupplots}
  \usetikzlibrary{calc}
  \usetikzlibrary{positioning}
  \usetikzlibrary{angles,quotes}
  \usetikzlibrary{backgrounds}
  \usetikzlibrary{external}
  \usetikzlibrary{fit}
  \usetikzlibrary{bayesnet}
  \usetikzlibrary{calc}
  \usetikzlibrary{fit}
  \usetikzlibrary{spy, shadows}
  \usetikzlibrary{arrows.meta}
  \usetikzlibrary{shadings}
  \usetikzlibrary{fadings}
\usetikzlibrary{matrix}     % SI units
\usepackage[
    noabbrev,
    capitalize,
    nameinlink]{cleveref}   % Automatic referencing

\begingroup
    \makeatletter
    \@for\theoremstyle:=theorem,corollary,lemma,model,definition,remark,plain\do{%
        \expandafter\g@addto@macro\csname th@\theoremstyle\endcsname{%
            \setlength\thm@preskip\parskip
            \setlength\thm@postskip\parskip%{0pt}
            \addtolength\thm@preskip\parskip
            }%
        }
    \makeatother
\endgroup

\usepackage{thmtools,thm-restate}

\theoremstyle{definition}

\newtheorem*{theorem*}{Theorem}

\newtheorem*{lemma*}{Lemma}

\newtheorem*{remark*}{Remark}

\newtheorem*{definition*}{Definition}

\newtheorem*{proposition*}{Proposition}

\newtheorem*{property*}{Property}

\crefname{thm}{Theorem}{Theorems}
\Crefname{thm}{Theorem}{Theorems}
\crefname{lem}{Lemma}{Lemmas}
\Crefname{lem}{Lemma}{Lemmas}
\crefname{prop}{Proposition}{Propositions}
\Crefname{prop}{Proposition}{Propositions}
\crefname{definition}{Definition}{Definitions}
\Crefname{definition}{Definition}{Definitions}
\crefname{property}{Property}{Properties}
\Crefname{property}{Property}{Properties}

\newcolumntype{L}{>{\RaggedRight\arraybackslash}X}

\definecolor{tabgreen}{HTML}{59a14f}

\newcommandx{\jbnote}[2][1=]{\todo[linecolor=red,backgroundcolor=red!25,bordercolor=red,#1]{JFB: #2}}
\newcommandx{\jrrnote}[2][1=]{\todo[linecolor=blue,backgroundcolor=blue!25,bordercolor=blue,#1]{JRR: #2}}
\newcommandx{\jgnote}[2][1=]{\todo[linecolor=green,backgroundcolor=green!25,bordercolor=green,#1]{JG: #2}}
\newcommandx{\retnote}[2][1=]{\todo[linecolor=yellow,backgroundcolor=yellow!25,bordercolor=yellow,#1]{RET: #2}}
\newcommandx{\thiswillnotshow}[2][1=]{\todo[disable,#1]{#2}}

\usepackage{microtype}
\usepackage{graphicx}
\usepackage{subcaption}
\usepackage{booktabs} % for professional tables
\usepackage{todonotes}

\usepackage{hyperref}

\usepackage[capitalize]{cleveref}

\usepackage{tikz}
\usetikzlibrary{decorations}
\usetikzlibrary{calc}
\usetikzlibrary{positioning}
\usetikzlibrary{angles,quotes}
\usetikzlibrary{backgrounds}
\usetikzlibrary{external}
\usetikzlibrary{fit}
\usetikzlibrary{bayesnet}
\usetikzlibrary{calc}
\usetikzlibrary{fit}
\usetikzlibrary{spy, shadows}
\usetikzlibrary{arrows.meta}
\usetikzlibrary{shadings}
\usetikzlibrary{fadings}
\usetikzlibrary{matrix}

\usepackage{standalone}
\usepackage{amsmath}
\usepackage{adjustbox}

\newcommand{\tasknorm}{\textsc{TaskNorm}}
\newcommand{\metanorm}{\textsc{MetaBN}}
\newcommand{\cnaps}{\textsc{CNAPs}}

\usepackage[accepted]{icml2020}

\icmltitlerunning{\tasknorm{}: Rethinking Batch Normalization for Meta-Learning}

\begin{document}

\twocolumn[
%\icmltitle{A Closer Look at Batch Normalization for Meta-Learning}
\icmltitle{\tasknorm{}: Rethinking Batch Normalization for Meta-Learning}



\icmlsetsymbol{equal}{*}

\begin{icmlauthorlist}
\icmlauthor{John Bronskill}{equal,cam}
\icmlauthor{Jonathan Gordon}{equal,cam}
\icmlauthor{James Requeima}{cam,inv}
\icmlauthor{Sebastian Nowozin}{ms}
\icmlauthor{Richard E. Turner}{cam,ms}
\end{icmlauthorlist}

\icmlaffiliation{cam}{University of Cambridge}
\icmlaffiliation{inv}{Invenia Labs}
\icmlaffiliation{ms}{Microsoft Research}

\icmlcorrespondingauthor{John Bronskill}{jfb54@cam.ac.uk}
\icmlkeywords{Meta-Learning, Few-shot Learning, Batch Normalization}

\vskip 0.3in
]



\printAffiliationsAndNotice{\icmlEqualContribution} % otherwise use the standard text.

\begin{abstract}
Modern meta-learning approaches for image classification rely on increasingly deep networks to achieve state-of-the-art performance, making batch normalization an essential component of meta-learning pipelines. 
However, the hierarchical nature of the meta-learning setting presents several challenges that can render conventional batch normalization ineffective, giving rise to the need to rethink normalization in this setting. 
We evaluate a range of approaches to batch normalization for meta-learning scenarios, and develop a novel approach that we call \tasknorm{}. 
Experiments on fourteen datasets demonstrate that the choice of batch normalization has a dramatic effect on both classification accuracy and training time for both gradient based- and gradient-free meta-learning approaches. 
Importantly, \tasknorm{} is found to consistently improve performance. 
Finally, we provide a set of best practices for normalization that will allow fair comparison of meta-learning algorithms.
\end{abstract}

\section{Introduction}

Meta-learning, or learning to learn \citep{thrun2012learning, schmidhuber1987evolutionary}, is an appealing approach for designing learning systems. 
It enables practitioners to construct models and training procedures that explicitly target desirable charateristics such as sample-efficiency and out-of-distribution generalization. 
Meta-learning systems have been demonstrated to excel at complex learning tasks such as few-shot learning  \citep{snell2017prototypical, finn2017model} and continual learning \citep{nagabandi2018deep, requeima2019cnaps, jerfel2019reconciling}.

Recent approaches to meta-learning rely on increasingly deep neural network based architectures to achieve state-of-the-art performance in a range of benchmark tasks \citep{finn2017model, mishra2018simple, triantafillou2019meta, requeima2019cnaps}. 
When constructing very deep networks, a standard component is the use of normalization layers (NL). 
In particular, in the image-classification domain, batch normalization \citep[BN;][]{ioffe2017batch} is crucial to the successful training of very deep convolutional networks.

However, as we discuss in \cref{sec:issues}, standard assumptions of the meta-learning scenario violate the assumptions of BN and vice-versa, complicating the deployment of BN in meta-learning. 
Many papers proposing novel meta-learning approaches employ different forms of BN for the proposed models, and some forms make implicit assumptions that, while improving benchmark performance, may result in potentially undesirable behaviours. 
Moreover, as we demonstrate in \cref{sec:experiments}, performance of the trained models can vary significantly based on the form of BN employed, confounding comparisons across methods. 
Further, naive adoption of BN for meta-learning does not reflect the statistical structure of the data-distribution in this scenario. 
In contrast, we propose a novel variant of BN -- \tasknorm{} -- that explicitly accounts for the statistical structure of the data distribution. 
We demonstrate that by doing so, \tasknorm{} further accelerates training of models using meta-learning while achieving improved test-time performance. 
Our main contributions are as follows:
\begin{itemize}[wide, labelwidth=!, labelindent=0pt]
\itemsep0em 
    \item We identify and highlight several issues with BN schemes used in the recent meta-learning literature.
    \item We propose \tasknorm{}, a novel variant of BN which is tailored for the meta-learning setting.
    \item In experiments with fourteen datasets, we demonstrate that \tasknorm{} consistently outperforms competing methods, while making less restrictive assumptions than its strongest competitor.
\end{itemize}

\section{Background and Related Work}

In this section we lay the necessary groundwork for our investigation of batch normalization in the meta-learning scenario.
Our focus in this work is on image classification. 
We denote images $\vx \in \sR^{C \times W \times H}$ where $W$ is the image width, $H$ the image height, $C$ the number of image channels. 
Each image is associated with a label $y \in \{1,\hdots, M\}$ where $M$ is the number of image classes. 
Finally, a dataset is denoted $D = \{(\vx_n, y_n)\}_{n=1}^N$. 

\subsection{Meta-Learning}
\label{sec:meta_learning}

We consider the meta-learning classification scenario. 
Rather than a single, large dataset $D$, we assume access to a dataset $\mathcal{D} = \{ \tau_t \}_{t = 1}^{K}$ comprising a large number of training \textit{tasks} $\tau_t$, drawn i.i.d. from a distribution $p(\tau)$. 
The data for a task $\tau$ consists of a \textit{context set} $D^\tau=\{(\vx^\tau_n, y^\tau_n)\}_{n=1}^{N_\tau}$ with $N_\tau$ elements with the inputs $\vx^\tau_n$ and labels $y^\tau_n$ observed, and a \textit{target set} $T^\tau=\{(\vx^{\tau\ast}_m, y^{\tau\ast}_m)\}_{m=1}^{M_\tau}$ with $M_\tau$ elements for which we wish to make predictions.
Here the inputs $\vx^{\tau\ast}$ are observed and the labels $y^{\tau\ast}$ are only observed during meta-training (i.e., training of the meta-learning algorithm).
The examples from a single task are assumed i.i.d., but examples across tasks are not.
Note that the target set examples are drawn from the same set of labels as the examples in the context set.

At meta-test time, the meta-learner is required to make predictions for target set inputs of unseen tasks. 
Often, the assumption is that test tasks will include classes that have not been seen during meta-training, and $D^{\tau}$ will contain only a few observations. 
The goal of the meta-learner is to process $D^{\tau}$, and produce a model that can make predictions for any test inputs $\vx^{\tau\ast} \in T^{\tau\ast}$ associated with the task.
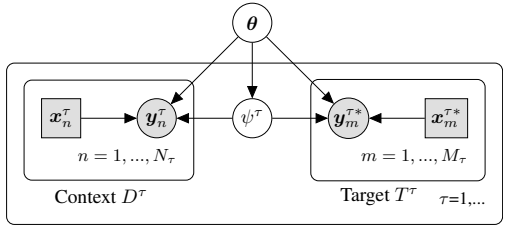
\begin{figure}
	\centering
	\begin{tikzpicture}
% nodes
\node[obs, rectangle, inner sep=.3em,draw] (x) {$\vx_m^{\tau\ast}$};%
\node[obs,left=of x] (y) {$\vy_{m}^{\tau\ast}$}; %
\node[latent, left=of y] (w) {$\psi^{\tau}$}; %
\node[obs, left=of w] (yt) {$\vy_n^{\tau}$}; %
\node[obs, rectangle, left=of yt, inner sep=.3em,draw] (xt) {$\vx_n^{\tau}$};%
\node[latent, above=of w, inner sep=.3em] (t) {$\vtheta$}; %
\node[draw=none,fill=none] at (-6.2, -1.4) {Context $D^{\tau}$}; 
\node[draw=none,fill=none] at (-1.2, -1.4) {Target $T^{\tau}$}; 
% plate
\plate [inner sep=.3cm, yshift=.02cm] {plate1} {(x) (y)} {$m=1,...,M_{\tau}$}; %
\plate [inner sep=.3cm, yshift=.02cm] {plate2} {(xt) (yt)} {$n=1,...,N_{\tau}$}; %
\plate [inner sep=.3cm, yshift=-.02cm] {plate3} {(plate1) (w) (plate2)} {$\tau$=1,...}; %

% underbrace for training plate
% \draw [thick, decoration={brace, mirror, raise=0.1cm
%     },
%     decorate
% ] (plate2.south west) -- (plate2.south east);

% edges
\edge {x} {y}; %
\edge {xt} {yt}; %
\edge {w} {y, yt}; %
\edge {t} {w, y, yt}; %

\end{tikzpicture}
	\caption[short]{Directed graphical model for multi-task meta-learning.}
	\label{fig:model_graph}
\end{figure}
\paragraph{Meta-Learning as Hierarchical Probabilistic Modelling}

A general and useful view of meta-learning is through the perspective of hierarchical probabilistic modelling \citep{heskes2000empirical, bakker2003task, grant2018recasting, gordon2018meta}. 
A standard graphical representation of this modelling approach is presented in \cref{fig:model_graph}. 
Global parameters $\vtheta$ encode information shared across all tasks, while local parameters $\vpsi^\tau$ encode information specific to task $\tau$.
This model introduces a \textit{hierarchy} of latent parameters, corresponding to the hierarchical nature of the data distribution. 

A general approach to meta-learning is to design inference procedures for the task-specific parameters $\vpsi^\tau = f_\vphi(D^\tau)$ conditioned on the context set \citep{grant2018recasting,gordon2018meta}, where $f$ is parameterized by additional parameters $\vphi$. 
Thus, a meta-learning algorithm defines a predictive distribution parameterized by $\vtheta$ and $\vphi$ as 
$
    p \left( y_m^{\tau\ast} | \vx_m^{\tau\ast}, f_\vphi \left( D^\tau \right), \vtheta \right).
$
This perspective relates to the \textit{inner} and \textit{outer} loops of meta-learning algorithms \citep{grant2018recasting, rajeswaran2019meta}: the inner loop uses $f_\vphi$ to provide local updates to $\vpsi$, while the outer loop provides predictions for target points.
Below, we use this view to summarize a range of meta-learning approaches.

\paragraph{Episodic Training}
\label{sec:episodic_training}

The majority of modern meta-learning methods employ \textit{episodic} training \citep{vinyals2016matching}. 
During meta-training, a task $\tau$ is drawn from $p(\tau)$ and randomly split into a context set $D^\tau$ and target set $T^\tau$.
The meta-learning algorithm's inner-loop is then applied to the context set to produce $\vpsi^\tau$.
With $\vtheta$ and $\vpsi^\tau$, the algorithm can produce predictions for the target set inputs $\vx_m^{\tau\ast}$. 

Given a differentiable loss function, and assuming that $f_\vphi$ is also differentiable, the meta-learning algorithm can then be trained with stochastic gradient descent algorithms.
Using log-likelihood as an example loss function, we may express a meta-learning objective for $\vtheta$ and $\vphi$ as
\begin{equation}
    \label{eqn:meta_learning_objective}
    \mathcal{L}(\vtheta, \vphi) = \mathop\mathbb{E}_{p(\tau)} \left[ \sum_{m=1}^{M_\tau} \log p \left( y_m^{\tau\ast} | \vx_m^{\tau\ast}, f_\vphi \left( D^\tau \right), \vtheta \right)  \right].
\end{equation}

\paragraph{Common Meta-Learning Algorithms}
\label{sec:meta_learning_algorithms}

There has been an explosion of meta-learning algorithms proposed in recent years. For an in-depth review see \citet{hospedales2020meta}. 
Here, we briefly introduce several methods, focusing on those that are relevant to our experiments. 
Arguably the most widely used is the gradient-based approach, the canonical example for modern systems being MAML \citep{finn2017model}.
MAML sets $\vtheta$ to be the initialization of the neural network parameters.
The local parameters $\vpsi^\tau$ are the network parameters after applying one or more gradient updates based on $D^\tau$.
Thus, $f$ in the case of MAML is a gradient-based procedure, which may or may not have additional parameters (e.g., learning rate).

Another widely used class of meta-learners are \textit{amortized-inference} based approaches e.g, \textsc{Versa}  \citep{gordon2018meta} and \textsc{CNAPs} \citep{requeima2019cnaps}. 
In these methods, $\vtheta$ parameterizes a shared feature extractor, and $\vpsi$ a set of parameters used to \textit{adapt} the network to the local task, which include a linear classifier and possibly additional parameters of the network.
For these models, $f$ is implemented via \textit{hyper-networks} \citep{ha2016hypernetworks} with parameters $\vphi$.
An important special case of this approach is Prototypical Networks (ProtoNets) \citep{snell2017prototypical}, which replace $\vpsi$ with nearest neighborhood classification in the embedding space of a learned feature extractor $g_\vtheta$. 
\subsection{Normalization Layers in Deep Learning}
\label{sec:batchnorm}

Normalization layers (NL) for deep neural networks were introduced by \citet{ioffe2015batch} to accelerate the training of neural networks by allowing the use of higher learning rates and decreasing the sensitivity to network initialization. 
Since their introduction, they have proven to be crucial components in the successful training of ever-deeper neural architectures.
Our focus is the few-shot image classification setting, and as such we concentrate on NLs for 2D convolutional networks. 
The input to a NL is $A = (\va_1, \hdots, \va_B)$, a batch of $B$ image-shaped activations or pre-activations, to which the NL is applied as
\begin{equation}
\label{eq:normalize}
    \va'_n = \vgamma \left( \frac{\va_n - \vmu} { \sqrt{ \vsigma^2 + \epsilon}} \right) + \vbeta, \quad 
\end{equation}
where $\vmu$ and $\vsigma$ are the \textit{normalization moments}, $\vgamma$ and $\vbeta$ are learned parameters, $\epsilon$ is a small scalar to prevent division by 0, and operations between vectors are element-wise.
NLs differ primarily by how the normalization moments are computed. 
The first such layer -- batch normalization (BN) -- was introduced by \citet{ioffe2015batch}.
A BN layer distinguishes between training and test modes.
At training time, BN computes the moments as
\begin{align}
    \vmu_{BN_c} &= \frac{1}{BHW} \sum_{b=1}^{B}  \sum_{w=1}^{W} \sum_{h=1}^{H} \va_{bwhc},\\
    \vsigma_{BN_c}^2 &= \frac{1}{BHW} \sum_{b=1}^{B}  \sum_{w=1}^{W} \sum_{h=1}^{H} (\va_{bwhc} - \vmu_{BN_c})^2.
\end{align}
Here, $\vmu_{BN}, \vsigma_{BN}^2, \vgamma, \vbeta \in \sR^C$.
As $\vmu_{BN}$ and $\vsigma_{BN}^2$ depend on the batch of observations, BN can be susceptible to failures if the batches at test time differ significantly from training batches, e.g., for streaming predictions.
To counteract this, at training time, a running mean and variance, $\vmu_r, \vsigma_r \in \sR^C$, are also computed for each BN layer over all training tasks and stored. 
At test time, test activations $\va$ are normalized using \cref{eq:normalize} with the statistics $\vmu_r$ and $\vsigma_r$ in place of the batch statistics.
Importantly, BN relies on the implicit assumption that $D$ comprises i.i.d. samples from some underlying distribution.

More recently, additional NLs have been introduced.
Many of these methods differ from standard BN in that they normalize each instance independently of the remaining instances in the batch, making them more resilient to batch distribution shifts at test time. 
These include instance normalization \citep{ulyanov2016instance}, layer normalization \citep{ba2016layer}, and group normalization \citep{wu2018group}. These are discussed further in \cref{sec:instance_based_norm}.

\subsection{Desiderata for Meta-Learning Normalization Layers}

As modern approaches to meta-learning systems routinely employ deep networks, NLs become essential for efficient training and optimal classification performance.
For BN in the standard supervised settings, i.i.d. assumptions about the data distribution imply that estimating moments from the training set will provide appropriate normalization statistics for test data. 
However, this does not hold in the meta-learning scenario, for which data points are only assumed to be i.i.d. within a specific task.
Therefore, the choice of what moments to use when applying a NL to the context and target set data points, during both meta-training and meta-test time, is key.

\begin{figure*}
	\centering
	\includegraphics[width=\textwidth]{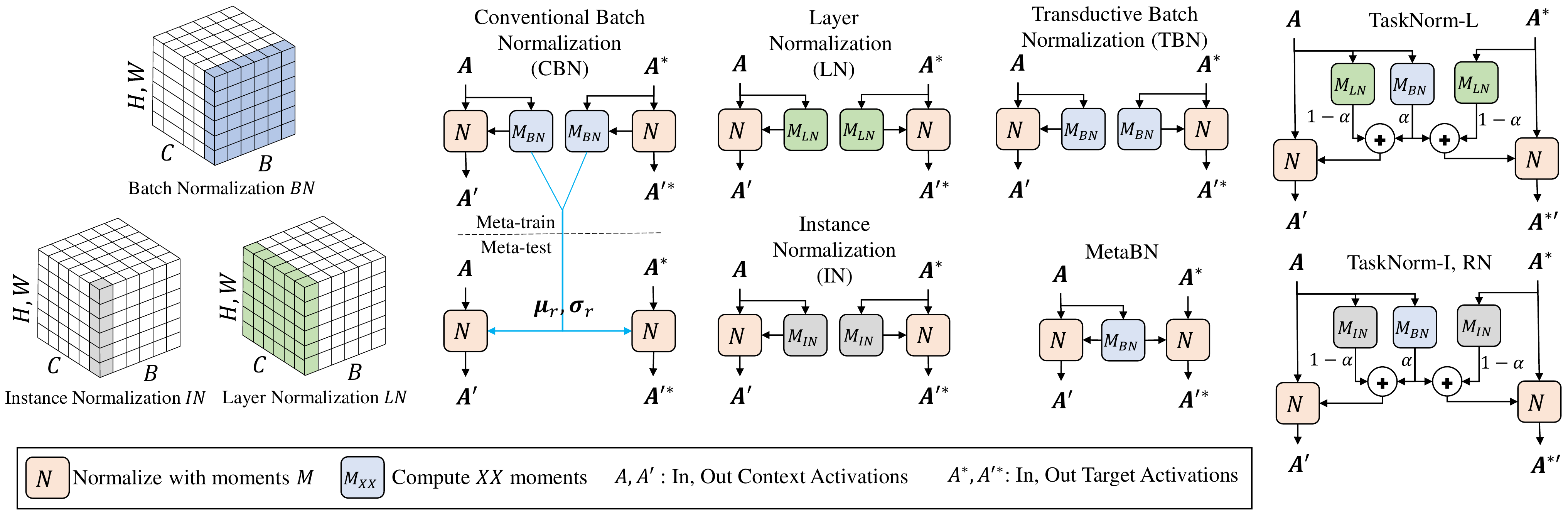}
	\caption[short]{A range of options for batch normalization for meta-learning. The cubes on the left depict the dimensions over which different moments are calculated for normalization of 2D convolutional layers. The computational diagrams on the right show how context and target activations are processed for various normalization methods. For all methods except conventional BN (CBN), the processing is identical at meta-train and meta-test time. Cube diagrams are derived from \citet{wu2018group}.}
	\label{fig:normalization_methods}
\end{figure*}

As a result, recent meta-learning approaches employ several normalization procedures that differ according to these design choices. 
A range of choices are summarized in \cref{fig:normalization_methods}.
As we discuss in \cref{sec:issues} and demonstrate with experimental results, some of these have implicit, undesirable assumptions which have significant impact on both predictive performance and training efficiency.
We argue that an appropriate NL for the meta-learning scenario requires consideration of the data-generating assumptions associated with the setting. 
In particular, we propose the following desiderata for a NL when used for meta-learning:
\begin{enumerate}[wide, labelwidth=!, labelindent=0pt]
\itemsep0em 
    \item Improves speed and stability of training without harming test performance (test set accuracy or log-likelihood);
    \item Works well across a range of context set sizes;
    \item Is non-transductive, thus supporting inference at meta-test time in a variety of circumstances.
\end{enumerate}

A non-transductive meta-learning system makes predictions for a single test set label conditioned only on a single input and the context set, while a transductive meta-learning system conditions on additional samples from the test set:
\begin{equation}
    \underbracket{p(y_i^{\tau\ast} | \vx_i^{\tau\ast}, D^\tau)}_{\smash{\text{non-transductive}}} \quad ; \quad  \underbracket{p(y_i^{\tau\ast} | \vx_{i=1:m}^{\tau\ast}, D^\tau)}_{\smash{\text{transductive}}}.
\end{equation}
We argue that there are two key issues with transductive meta-learners. 
The first is that transductive learning is sensitive to the distribution over the target set used during meta-training, and as such is less generally applicable than non-transductive learning. 
For example, transductive learners may fail to make good predictions if target sets contain a different class balance than what was observed during meta-training, or if they are required to make predictions for one example at a time.
Transductive learners can also violate privacy constraints.
In \cref{tab:maml_results} and \cref{app:transduction_issues}, we provide empirical demonstrations of these failure cases.

The second issue is that transductive learners have more information available than non-transductive learners at prediction time, which may lead to unfair comparisons.
It is worth noting that some meta-learning algorithms are specifically designed to leverage transductive inference \citep[e.g.,][]{ren2018metalearning,liu2018learning}, though we do not discuss them in this work. 
In \cref{sec:experiments} we demonstrate that there are significant performance differences for a model when trained transductively versus non-transductively.

\section{Normalization Layers for Meta-learning}
\label{sec:issues}

In this section, we discuss several normalization schemes that can and have been applied in the recent meta-learning literature, highlighting the modelling assumptions and effects of different design choices. 
Throughout, we assume that the meta-learning algorithm is constructed such that the context-set inputs are passed through every neural-network module that the target set inputs are passed through at prediction time.
This implies that moments are readily available from both the context and target set observations for any normalization layer, and is the case for many widely-used meta-learning models \citep[e.g.,][]{finn2017model,snell2017prototypical,gordon2018meta}.

To illustrate our arguments, we provide experiments with \textsc{MAML} running simple, but widely used few-shot learning tasks from the Omniglot \citep{lake2011one} and miniImagenet \citep{ravi2016optimization} datasets. 
The results of these experiments are provided in \cref{tab:maml_results}, and full experimental details in \cref{app:experiment_details}.

\subsection{Conventional Usage of Batch Normalization (CBN)}
\label{sec:conventional_batchnorm}

We refer to \textit{conventional} batch normalization (CBN) as that defined by \citet{ioffe2015batch} and as outlined in \cref{sec:batchnorm}.
In the context of meta-learning, this involves normalizing tasks with computed moments at meta-train time, and using the accumulated running moments to normalize the tasks at meta-test time (see CBN in \cref{fig:normalization_methods}).

We highlight two important issues with the use of CBN for meta-learning. The first is that, from the graphical model perspective, this is equivalent to lumping $\vmu$ and $\vsigma$ with the global parameters $\vtheta$, i.e., they are learned from the meta-training set and shared across all tasks at meta-test time.
We might expect CBN to perform poorly in meta-learning applications since the running moments are \textit{global} across all tasks while the task data is only i.i.d. \textit{locally} within a task, i.e., CBN does not satisfy desiderata 1.
This is corroborated by our results (\cref{tab:maml_results}), where we demonstrate that using CBN with \textsc{MAML} results in very poor predictive performance - no better than chance. 
The second issue is that, as demonstrated by \citet{wu2018group}, using small batch sizes leads to inaccurate moments, resulting in significant increases in model error. 
Importantly, the small batch setting may occur often in meta-learning, for example in the 1-shot scenario.
Thus, CBN does not satisfy desiderata 2. 

Despite these issues, CBN is sometimes used, e.g., by \citet{snell2017prototypical}, though testing was performed only on Omniglot and miniImagenet where the distribution of tasks is homogeneous \citep{triantafillou2019meta}.
In \cref{sec:experiments}, we show that Batch renormalization \citep[BRN;][]{ioffe2017batch} can exhibit poor predictive performance in meta-learning scenarios (see \cref{app:batch_renorm} for further details).  

\subsection{Transductive Batch Normalization (TBN)}
\label{sec:transductive_batchnorm}

Another approach is to do away with the running moments used for normalization at meta-test time, and replace these with context / target set statistics. Here, context / target set statistics are used for normalization, both at meta-train \textit{and} meta-test time.
This is the approach taken by the authors of MAML \citep{finn2017model},\footnote{See for example \citep{Finn2017code} for a reference implementation.} and, as demonstrated in our experiments, seems to be crucial to achieve the reported performance.
From the graphical model perspective, this implies associating the normalization statistics with neither $\vtheta$ nor $\vpsi$, but rather with a special set of parameters that is local for each set (i.e., normalization statistics for $T^\tau$ are independent of $D^\tau$).
We refer to this approach as \textit{transductive} batch normalization (TBN; see \cref{fig:normalization_methods}).

Unsurprisingly, \citet{nichol2018first} found that using TBN provides a significant performance boost in all cases they tested, which is corroborated by our results in \cref{tab:maml_results}. 
In other words, TBN achieves desiderata 2, and, as we demonstrate in \cref{sec:experiments}, desiderata 1 as well. 
However, it is transductive.
Due to the ubiquity of MAML, many competetive meta-learning methods \citep[e.g.][]{gordon2018meta} have adopted TBN. 
However, in the case of TBN, transductivity is rarely stated as an explicit assumption, and may often confound the comparison among methods \citep{nichol2018first}.
Importantly, we argue that to ensure comparisons in experimental papers are rigorous, meta-learning methods that are transductive must be labeled as such.

\subsection{Instance-Based Normalization Schemes}
\label{sec:instance_based_norm}

An additional class of non-transductive NLs are \textit{instance-based} NLs. 
Here, both at meta-train and meta-test time, moments are computed separately for each instance, and do not depend on other observations.
From a modelling perspective, this corresponds to treating $\vmu$ and $\vsigma$ as local at the \textit{observation} level.
As instance-based NLs do not depend on the context set size, they perform equally well across context-set sizes (desiderata 2).
However, as we demonstrate in \cref{sec:experiments}, the improvements in predictive performance are modest compared to more suitable NLs and they are worse than CBN in terms of training efficiency (thus not meeting desiderata 1).
Below, we discuss two examples, with a third discussed in \cref{app:group_norm}.

\paragraph{Layer Normalization \citep[LN;][]{ba2016layer}}
LN (see \cref{fig:normalization_methods}) has been shown to improve performance compared to CBN in recurrent neural networks, but does not offer the same gains for convolutional neural networks \citep{ba2016layer}.
The LN moments are computed as:
\begin{align}
    {\vmu}_{LN_b} &= \frac{1}{HWC} \sum_{w=1}^{W} \sum_{h=1}^{H} \sum_{c=1}^{C}  \va_{bwhc},\\
    {\vsigma}_{LN_b}^2 &= \frac{1}{HWC} \sum_{w=1}^{W} \sum_{h=1}^{H} \sum_{c=1}^{C} (\va_{bwhc} - \vmu_{LN_b})^2
\end{align}
where $\vmu_{LN}, \vsigma_{LN}^2 \in \sR^B$.
While non-transductive, \cref{tab:maml_results} demonstrates that LN  falls far short of TBN in terms of accuracy. 
Further, in \cref{sec:experiments} we demonstrate that LN lacks in training efficiency when compared to other NLs.
\paragraph{Instance Normalization \citep[IN;][]{ulyanov2016instance}}
IN (see \cref{fig:normalization_methods}) has been used in a wide variety of image generation applications. 
The IN moments are computed as:
\begin{align}
    {\vmu}_{IN_{bc}} &= \frac{1}{HW} \sum_{w=1}^{W} \sum_{h=1}^{H} \va_{bwhc},\\
    {\vsigma}_{IN_{bc}}^2 &= \frac{1}{HW} \sum_{w=1}^{W} \sum_{h=1}^{H} (\va_{bwhc} - \vmu_{IN_{bc}})^2
\end{align}
where $\vmu_{IN}, \vsigma_{IN}^2 \in \sR^{B \times C}$.
\cref{tab:maml_results} demonstrates that IN has superior predictive performance to that of LN, but falls considerably short of TBN.
In \cref{sec:experiments} we show that IN lacks in training efficiency when compared to other NLs.

\section{Task Normalization}
\label{sec:task_normalization}

In the previous section, we demonstrated that it is not immediately obvious how NLs should be designed for meta-learning applications. 
We now develop \tasknorm{}, the first NL that is specifically tailored towards this scenario.
\tasknorm{} is motivated by the view of meta-learning as hierarchical probabilistic modelling, discussed in \cref{sec:meta_learning}. 
Given this hierarchical view of the model parameters, the question that arises is, how should we treat the normalization statistics $\vmu$ and $\vsigma$? 
\cref{fig:model_graph} implies that the data associated with a task $\tau$ are i.i.d. \textit{only when conditioning on both $\vtheta$ and $\vpsi^\tau$.} 
Thus, the normalization statistics $\vmu$ and $\vsigma$ should be local at the task level, i.e., absorbed into $\vpsi^\tau$. 
Further, the view that $\vpsi^\tau$ should be inferred conditioned on $D^\tau$ implies that the normalization statistics for the target set should be computed directly from the context set. 
Finally, our desire for a non-transductive scheme implies that any contribution from points in the target should not affect the normalization for other points in the target set, i.e., when computing $\vmu$ and $\vsigma$ for a particular observation $\vx^{\tau\ast} \in T^\tau$, the NL should only have access to $D^\tau$ and $\vx^{\tau\ast}$.
\subsection{Meta-Batch Normalization (\metanorm)}
This perspective leads to our definition of \metanorm{}, which is a simple adaptation of CBN for the meta-learning setting. 
In \metanorm{}, the context set alone is used to compute the normalization  statistics for \textit{both} the context and target sets, both at meta-train and meta-test time (see \cref{fig:normalization_methods}). To our knowledge, \metanorm{} has not been described in any publication, but concurrent to this work, it is used in the implementation of Meta-Dataset \citep{Triantafillou2019code}.

\metanorm{} meets almost all of our desiderata, it 
\begin{inlinelist}
    \item is non-transductive since the normalization of a test input does not depend on other test inputs in the target set, and
    \item as we demonstrate in \cref{sec:experiments}, it improves training speed while maintaining accuracy levels of meta-learning models.
\end{inlinelist}
However, as we demonstrate in \cref{sec:experiments}, \metanorm{} performs less well for small context sets.
This is because moment estimates will have high-variance when there is little data, and is similar to the difficulty of using BN with small-batch training \citep{wu2018group}.
To address this issue, we introduce the following extension to \metanorm{}, which yields our proposed normalization scheme -- \tasknorm{}.

\subsection{\tasknorm{}}
\label{sec:tasknorm}

The key intuition behind \tasknorm{} is to normalize a task with the context set moments in combination with a set of non-transductive, secondary moments computed from the input being normalized. 
A blending factor $\alpha$ between the two sets of moments is learned during meta-training. 
The motivation for \tasknorm{} is as follows: when the context set $D^\tau$ is small (e.g. 1-shot or few-shot learning) the context set alone will lead to noisy and inaccurate estimates of the ``true'' task statistics.
In such cases, a secondary set of moments may improve the estimate of the moments, leading to better training efficiency and predictive performance in the low data regime.
Further, this provides information regarding $\vx^{\tau\ast}$ at prediction time while maintaining non-transductivity.
The pooled moments for \tasknorm{} are computed as:
\begin{align}
\label{eq:blending}
    \vmu_{TN} =  & \alpha \vmu_{BN} + (1-\alpha)\vmu_+,\\
    \vsigma_{TN}^2 = & \alpha\left( \vsigma_{BN}^2 + (\vmu_{BN} - \vmu_{TN})^2 \right) \nonumber\\ 
    & + (1 - \alpha) \left(\vsigma_+^2 + (\vmu_+ - \vmu_{TN})^2 \right), \label{eqn:tn_sigma}
\end{align}

where $\vmu_{TN}, \vsigma_{TN} \in \sR^{B \times C}$, $\vmu_+$, $\vsigma_+^2$ are additional moments from a non-transductive NL such as LN or IN computed using activations from the example being normalized (see \cref{fig:normalization_methods}), and $\vmu_{BN}$ and $\vsigma_{BN}$ are computed from $D^\tau$.
\cref{eqn:tn_sigma} is the standard \textit{pooled variance} when combining the variance of two Gaussian estimators.

\begin{figure}
\centering
    \begin{subfigure}{0.7\columnwidth}
    \centering
    \includegraphics[width=\columnwidth]{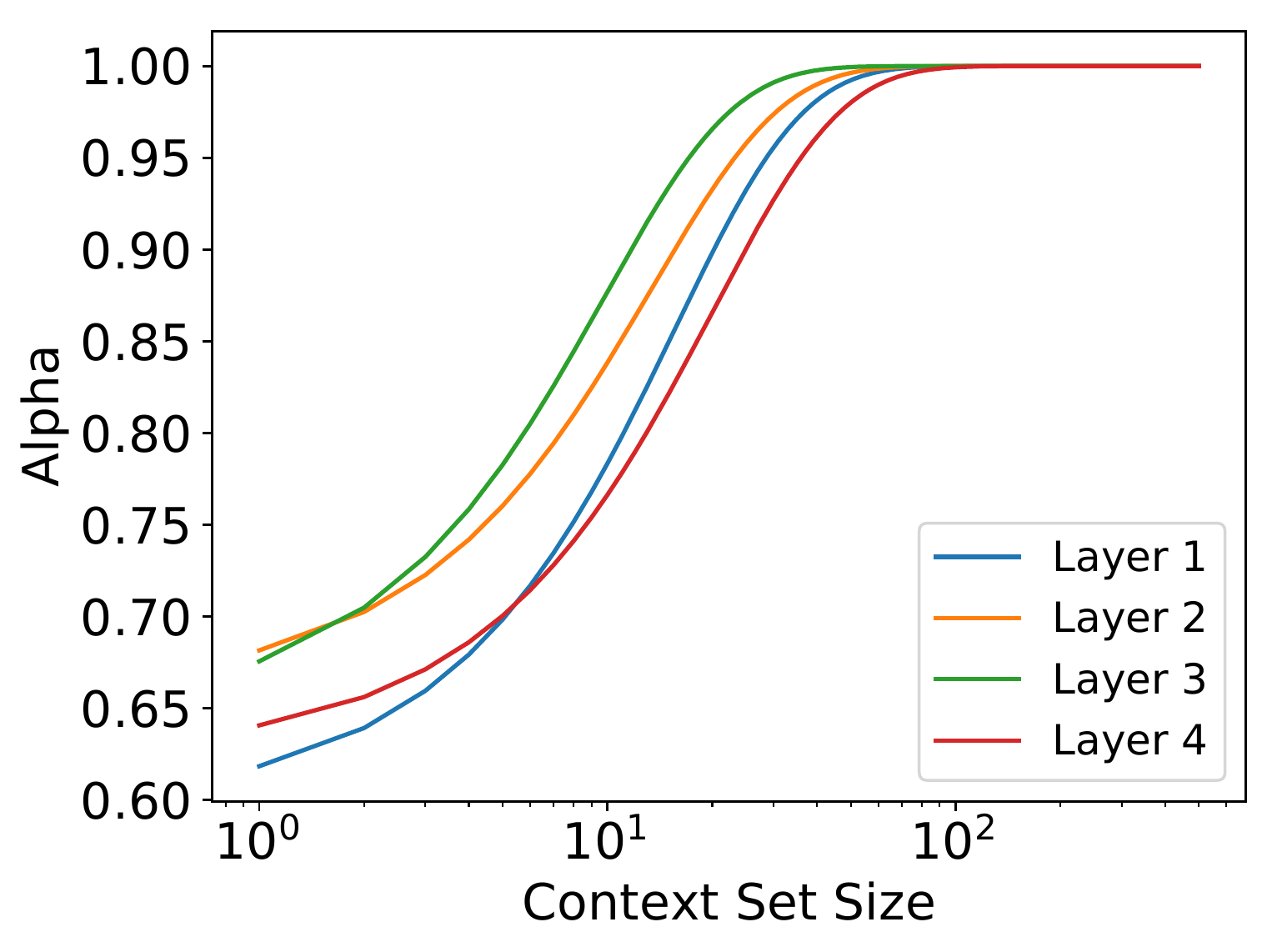}
    \caption{}
    \label{fig:alpha_curves}
    \end{subfigure}%
    \hfill
    \begin{subfigure}{0.7\columnwidth}
    \centering
    \includegraphics[width=\columnwidth]{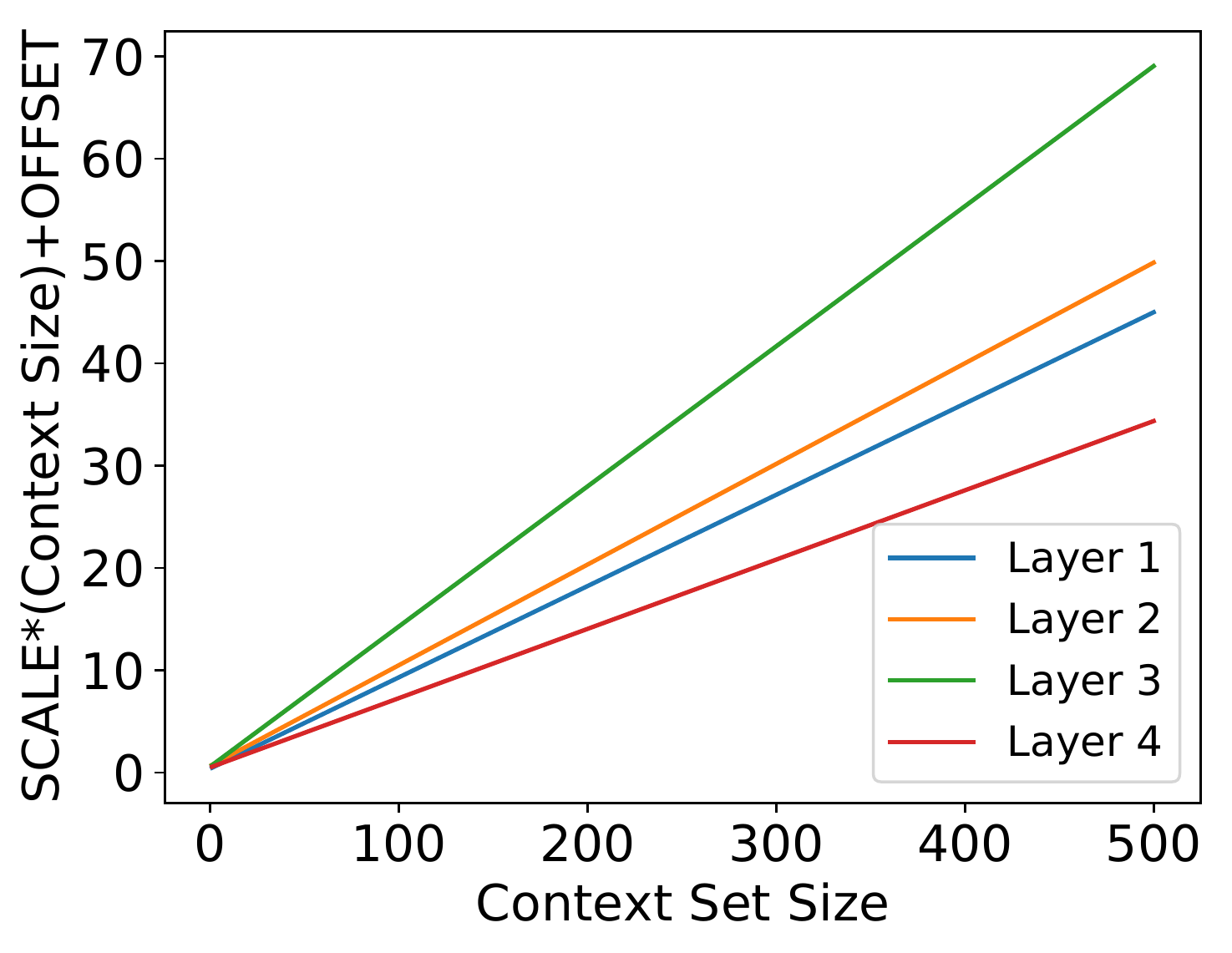}
    \caption{}
    \label{fig:a_b_curves}
    \end{subfigure}
    \caption{Plots of: (a) $\alpha$ versus context set size, and (b) $\alpha$ versus $\textsc{SCALE} |D^{\tau}| + \textsc{OFFSET}$ for the first NL in each of the four layers in the feature extractor for the \tasknorm{}-I model.}
\end{figure}

Importantly, we parameterize $\alpha = \textsc{sigmoid}(\textsc{scale} |D^{\tau}| + \textsc{offset})$, where the \textsc{sigmoid} function ensures that $0 \leq \alpha \leq 1$, and the scalars \textsc{scale} and \textsc{offset} are learned during meta-training.
This enables us to learn how much each set should contribute to the estimate of task statistics as a function of the context-set size $|D^\tau|$.
\cref{fig:alpha_curves} depicts the value of $\alpha$ as a function of context set size $|D^{\tau}|$ for a representative set of trained \tasknorm{} layers.
In general, when the context size is suitably large ($N_\tau > 25$), $\alpha$ is close to unity, i.e., normalization is carried out entirely with the context set in those layers. 
When the context size is smaller, there is a mix of the two sets of moments.

Allowing each \tasknorm{} layer to separately adapt to the size of the context set (as opposed to learning a fixed $\alpha$ per layer) is crucial in the meta-learning setting, where we expect the size of $D^\tau$ to vary, and are often particularly interested in the ``few-shot'' regime.
\cref{fig:a_b_curves} plots the line $\textsc{SCALE} |D^{\tau}| + \textsc{OFFSET}$ for same set of NLs as \cref{fig:alpha_curves}. 
The algorithm has learned that the \textsc{SCALE} parameter is non-zero and the \textsc{OFFSET} is almost zero in all cases indicating the importance of having $\alpha$ being a function of context size.
In \cref{app:ablation_study}, we provide an ablation study demonstrating the importance of our proposed parameterization of $\alpha$.
If the context size is fixed, we do not use the full parameterization, but learn a single value for alpha directly.
The computational cost of \tasknorm{} is marginally greater than CBN's. As a result, per-iteration time increases only slightly. However, as we show in \cref{sec:experiments}, \tasknorm{} converges faster than CBN.

In related work, \citet{nam2018batch} define Batch-Instance Normalization (BIN) that combines the results of CBN and IN with a learned blending factor in order to attenuate unnecessary styles from images. However, BIN blends the output of the individual CBN and IN normalization operations as opposed to blending the moments.

Finally, we note that Reptile \citep{nichol2018first} uses a non-transductive form of task normalization that involves normalizing examples from the target set one example at a time with the moments of the context set augmented with the single example.
We refer to this approach as \textit{reptile} normalization or RN. 
It is easy to show that RN is a special case of \tasknorm{} augmented with IN when %
$
    \alpha = |D^{\tau}| / (1 + |D^{\tau}|).
$
In \cref{sec:experiments}, we show that reptile normalization falls short of \tasknorm{}, supporting the intuition that learning the value of $\alpha$ is preferable to fixing a value.

\section{Experiments}
\label{sec:experiments}

In this section, we evaluate \tasknorm{} along with a range of competitive normalization approaches.\footnote{Source code is available at \url{https://github.com/cambridge-mlg/cnaps}}
The goal of the experiments is to evaluate the following hypotheses:
\begin{inlinelist}
    \item Meta-learning algorithms are sensitive to the choice of NL;
    \item TBN will, in general, outperform non-transductive NLs; and
    \item NLs that consider the meta-learning data assumptions (\tasknorm{}, \metanorm{}, RN) will outperform ones that do not (CBN, BRN, IN, LN, etc.).
\end{inlinelist}
\subsection{Small Scale Few-Shot Classification Experiments}
\label{sec:small_scale_experiments}

\begin{table*}[t]
\caption{Accuracy results for different few-shot settings on Omniglot and miniImageNet using the MAML algorithm. All figures are percentages and the $\pm$ sign indicates the 95\% confidence interval. Bold indicates the highest scores. The numbers after the configuration name indicate the way and shots, respectively. The vertical lines enclose the transductive results. The \textit{TBN}, \textit{examples}, and \textit{class} columns indicate accuracy when tested with all target examples at once, one example at a time, and one class at a time, respectively. All other NLs are non-transductive and yield the same result when tested by example or class.}
\label{tab:maml_results}
%\vskip 0.1in
\begin{center}
\begin{small}
\begin{adjustbox}{max width=\textwidth}
\begin{tabular}{ l |c c c| c c c c c c c c}
\toprule
Configuration     & TBN                   & example & class & CBN          & BRN           & LN           & IN           & RN                    & MetaBN                & TaskNorm-L            & TaskNorm-I \\
\midrule
Omniglot-5-1      & \textbf{98.4$\pm$0.7} & 21.6$\pm$1.3 & 21.6$\pm$1.3 & 20.1$\pm$0.0 & 20.0$\pm$0.0  & 83.0$\pm$1.3 & 87.4$\pm$1.2 & 92.6$\pm$0.9          & 91.8$\pm$0.9          & 94.0$\pm$0.8 & 94.4$\pm$0.8 \\ 
Omniglot-5-5      & \textbf{99.2$\pm$0.2} & 22.0$\pm$0.5 & 23.2$\pm$0.5 & 20.0$\pm$0.0 & 20.0$\pm$0.0  & 91.0$\pm$0.8 & 93.9$\pm$0.5 & 98.2$\pm$0.2          & 98.1$\pm$0.3          & 98.0$\pm$0.3 & 98.6$\pm$0.2 \\ 
Omniglot-20-1     & \textbf{90.9$\pm$0.5} &  3.7$\pm$0.2 &  3.7$\pm$0.2 & 5.0$\pm$0.0  &  5.0$\pm$0.0  & 78.1$\pm$0.7 & 80.4$\pm$0.7 & 89.0$\pm$0.6          & 89.6$\pm$0.5          & 89.6$\pm$0.5 & \textbf{90.0$\pm$0.5} \\ 
Omniglot-20-5     & \textbf{96.6$\pm$0.2} &  5.5$\pm$0.2 & 14.5$\pm$0.3 & 5.0$\pm$0.0  &  5.0$\pm$0.0  & 92.3$\pm$0.2 & 92.9$\pm$0.2 & \textbf{96.8$\pm$0.2} & 96.4$\pm$0.2          & 96.4$\pm$0.2 & 96.3$\pm$0.2 \\ 
miniImageNet-5-1  & \textbf{45.5$\pm$1.8} & 26.9$\pm$1.5 & 26.9$\pm$1.5 & 20.1$\pm$0.0 & 20.4$\pm$0.4  & 41.2$\pm$1.6 & 40.7$\pm$1.7 & 40.7$\pm$1.7          & 41.6$\pm$1.6          & 42.0$\pm$1.7 & 42.4$\pm$1.7 \\ 
miniImageNet-5-5  & \textbf{59.7$\pm$0.9} & 30.3$\pm$0.7 & 27.2$\pm$0.6 & 20.2$\pm$0.2 & 20.7$\pm$0.5  & 52.8$\pm$0.9 & 54.3$\pm$0.9 & 57.6$\pm$0.9          & \textbf{58.6$\pm$0.9} & 58.1$\pm$0.9 & \textbf{58.7$\pm$0.9} \\
\midrule
Average Rank      & 1.25                  & -            & -            & 8.42         & 8.58          & 6.58         & 5.75         & 4.00                  & 3.67                  & 3.75        & 3.00 \\
\bottomrule
\end{tabular}
\end{adjustbox}
\end{small}
\end{center}
\end{table*}
\begin{figure*}
\centering
    \begin{subfigure}{0.33\textwidth}
    \centering
    \includegraphics[width=\columnwidth]{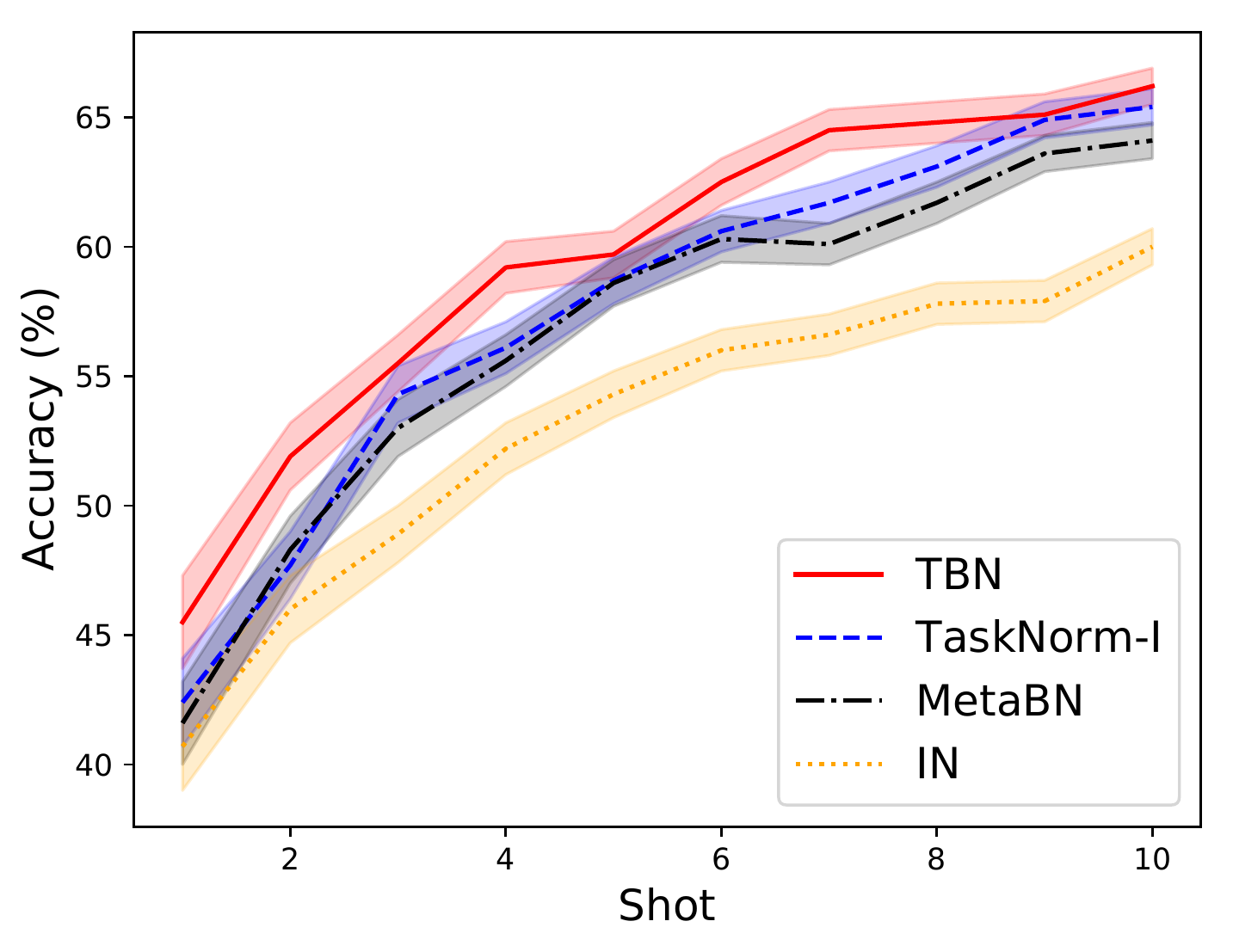}
    \caption{}
    \label{fig:maml_accuracy_vs_shot}
    \end{subfigure}%
    \hfill
    \begin{subfigure}{0.33\textwidth}
    \centering
    \includegraphics[width=\columnwidth]{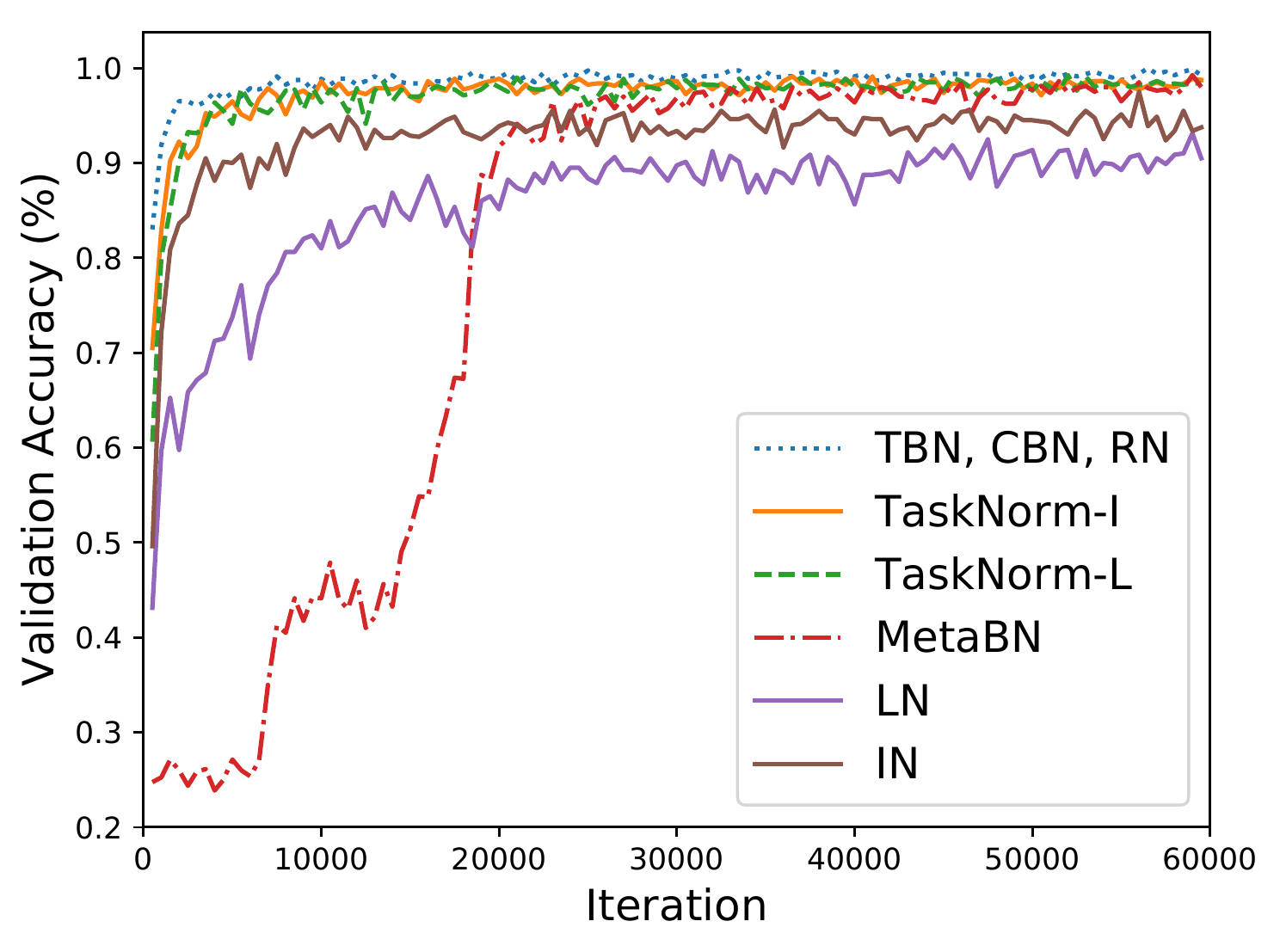}
    \caption{}
    \label{fig:omniglot_5_5_training_curves}
    \end{subfigure}%
    \hfill\begin{subfigure}{0.33\textwidth}
    \centering
    \includegraphics[width=\columnwidth]{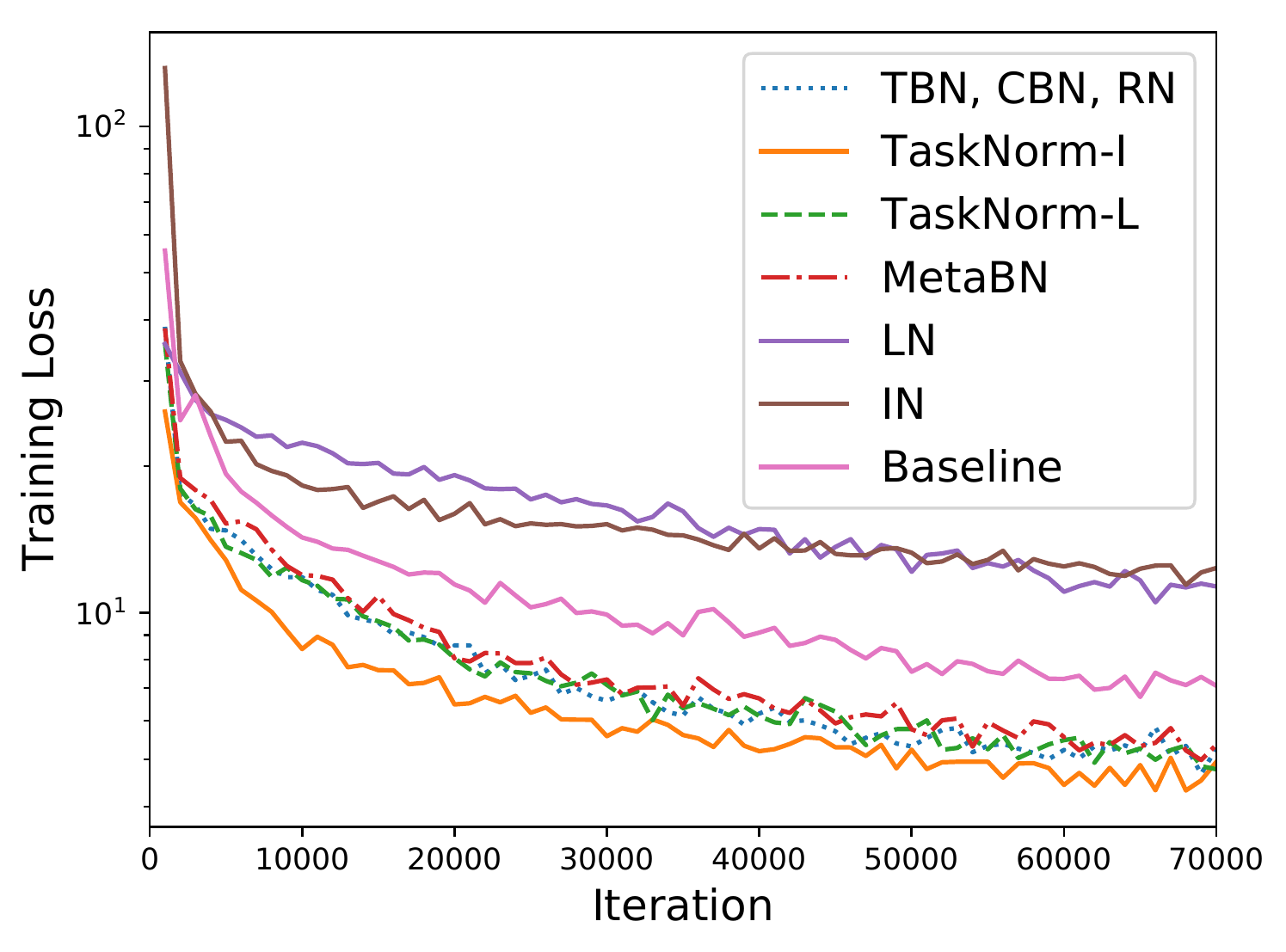}
    \caption{}
    \label{fig:cnaps_training_curves}
    \end{subfigure}
    \caption{(a) Accuracy vs shot for MAML on 5-way miniImagenet classification. (b) Plot of validation accuracy versus training iteration using MAML for Omniglot 5-way, 5-shot corresponding to the results in \cref{tab:maml_results}. (c) Training Loss versus iteration corresponding to the results using the CNAPS algorithm in \cref{tab:cnaps_meta_dataset}. Note that TBN, CBN, and RN all share the same meta-training step.}
\end{figure*}
We evaluate \tasknorm{} and a set of NLs using the first order MAML and ProtoNets algorithms on the Omniglot and miniImageNet datasets under various way (the number of classes used in each task) and shot (the number of context set examples used per class) configurations. 
This setting is smaller scale, and considers only fixed-sized context and target sets.
Configuration and training details can be found in \cref{app:experiment_details}.
\paragraph{Accuracy}
\cref{tab:maml_results} and \cref{tab:protonets_results} show accuracy results for various normalization methods on the Omniglot and miniImageNet datasets using the first order MAML and the ProtoNets algorithms, respectively.
We compute the average rank in an identical manner to \citet{triantafillou2019meta}. 

For MAML, TBN is clearly the best method in terms of classification accuracy. 
The best non-transductive approach is \tasknorm{} that uses IN augmentation (\tasknorm{}-I). 
The two methods using instance-based normalization (LN, IN) do significantly less well than methods designed with meta-learning desiderata in mind (i.e. \tasknorm{}, MetaBN, and RN). 
The methods using running averages at meta-test time (CBN, BRN) fare the worst.
\cref{fig:maml_accuracy_vs_shot} compares the performance of MAML on unseen tasks from miniImageNet when trained with TBN, IN, \metanorm{}, and \tasknorm{}, as a function of the number of shots per class in $D^\tau$, and demonstrates that these trends are consistent across the low-shot range.

Note that when meta-testing occurs one example at a time (e.g.~in the streaming data scenario) or one class at a time (unbalanced class distribution scenario), accuracy for TBN drops dramatically compared to the case where all the examples are tested at once. This is an important drawback of the transductive approach. All of the other NLs in the table are non-transductive and do not suffer a decrease in accuracy when tested an example at a time or a class at a time.

Compared to MAML, the ProtoNets algorithm is much less sensitive to the NL used. \cref{tab:protonets_results} indicates that with the exception of IN, all of the normalization methods yield good performance. We suspect that this is due to the fact that in ProtoNets employs a parameter-less nearest neighbor classifier and no gradient steps are taken at meta-test time, reducing the importance of normalization. The top performer is LN which narrowly edges out TaskNorm-L and CBN. Interestingly, TBN is not on top and \tasknorm{}-I lags as IN is the least effective method.

\paragraph{Training Speed} 
\cref{fig:omniglot_5_5_training_curves} plots validation accuracy versus training iteration for the first order MAML algorithm training on Omniglot 5-way-5-shot. 
TBN is the most efficient in terms of training convergence. 
The best non-transductive method is again \tasknorm{}-I, which is only marginally worse than TBN and just slightly better than \tasknorm{}-L. 
Importantly, \tasknorm{}-I is superior to either of MetaBN and IN alone in terms of training efficiency.
\cref{fig:proto_nets_o-20_1_training_curves} depicts the training curves for the ProtoNets algorithm. With the exception of IN which converges to a lower validation accuracy, all NLs converge at the the same speed.

For the MAML algorithm, the experimental results support our hypotheses. 
Performance varies significantly across NLs. 
TBN outperformed all methods in terms of classification accuracy and training efficiency, and \tasknorm{} is the best non-transductive approach. 
Finally, The meta-learning specific methods outperformed the more general ones. The picture for ProtoNets is rather different. There is little variability across NLs, TBN lagged the most consistent method LN in terms of accuracy, and the NLs that considered meta-learning needs were not necessarily superior to those that did not.

\subsection{Large Scale Few-Shot Classification Experiments}
\label{sec:large_scale_experiments}
\begin{table*}[h]

\caption{Few-shot classification results on \textsc{Meta-Dataset} using the \cnaps{} (top) and ProtoNets (bottom) algorithms. Meta-training performed on datasets above the dashed line. Datasets below the dashed line are entirely held out. All figures are percentages and the $\pm$ sign indicates the 95\% confidence interval over tasks. Bold indicates the highest scores. Vertical lines in the TBN column indicate that this method is transductive. Numbers in the \textsc{Baseline} column are from \citep{requeima2019cnaps}.}
\label{tab:cnaps_meta_dataset}
%\vskip 0.1in
\begin{center}
\begin{small}
\begin{adjustbox}{max width=\textwidth}
\begin{tabular}{l | c | c c c c c c c c c c}
\toprule
Dataset       & TBN                   & Baseline              & CBN          & BRN          & LN           & IN                    & RN                    & MetaBN                & TaskNorm-r   & TaskNorm-L            & TaskNorm-I \\
\midrule
ILSVRC        & \textbf{50.2$\pm$1.0} & \textbf{51.3$\pm$1.0} & 24.8$\pm$0.7 & 19.2$\pm$0.7 & 45.5$\pm$1.1 & 46.7$\pm$1.0          & 49.7$\pm$1.1          & \textbf{51.3$\pm$1.1} & 49.3$\pm$1.0          & \textbf{51.2$\pm$1.1} & \textbf{50.6$\pm$1.1} \\ 
Omniglot      & \textbf{91.4$\pm$0.5} & 88.0$\pm$0.7          & 47.9$\pm$1.4 & 60.0$\pm$1.6 & 87.4$\pm$0.8 & 79.7$\pm$1.0          & \textbf{91.0$\pm$0.6} & \textbf{90.9$\pm$0.6} & 87.8$\pm$0.7          & 90.6$\pm$0.6          & \textbf{90.7$\pm$0.6} \\
Aircraft      & 81.6$\pm$0.6          & 76.8$\pm$0.8          & 29.5$\pm$0.9 & 56.3$\pm$0.8 & 76.5$\pm$0.8 & 74.7$\pm$0.7          & 82.4$\pm$0.6          & \textbf{83.9$\pm$0.6} & 81.1$\pm$0.7          & 81.9$\pm$0.6          & \textbf{83.8$\pm$0.6} \\
Birds         & \textbf{74.5$\pm$0.8} & 71.4$\pm$0.9          & 42.1$\pm$1.0 & 32.6$\pm$0.8 & 67.3$\pm$0.9 & 64.9$\pm$1.0          & 72.4$\pm$0.8          & 73.2$\pm$0.9          & 72.8$\pm$0.9          & 72.4$\pm$0.8          & \textbf{74.6$\pm$0.8} \\
Textures      & 59.7$\pm$0.7          & \textbf{62.5$\pm$0.7} & 37.5$\pm$0.7 & 50.5$\pm$0.6 & 60.1$\pm$0.6 & 59.7$\pm$0.7          & 58.6$\pm$0.7          & 58.9$\pm$0.8          & \textbf{63.2$\pm$0.8} & 57.2$\pm$0.7          & 62.1$\pm$0.7 \\
Quick Draw    & 70.8$\pm$0.8          & 71.9$\pm$0.8          & 44.5$\pm$1.0 & 56.7$\pm$1.0 & 71.6$\pm$0.8 & 68.2$\pm$0.9          & \textbf{74.3$\pm$0.8} & \textbf{74.1$\pm$0.7} & 71.6$\pm$0.8          & \textbf{74.3$\pm$0.8} & \textbf{74.8$\pm$0.7} \\
Fungi         & 46.0$\pm$1.0          & 46.0$\pm$1.1          & 21.1$\pm$0.8 & 26.1$\pm$0.9 & 39.6$\pm$1.0 & 37.8$\pm$1.0          & \textbf{49.0$\pm$1.0} & \textbf{47.9$\pm$1.0} & 42.0$\pm$1.1          & 47.1$\pm$1.1          & \textbf{48.7$\pm$1.0} \\
VGG Flower    & 86.6$\pm$0.5          & \textbf{89.2$\pm$0.5} & 79.0$\pm$0.7 & 75.7$\pm$0.7 & 84.4$\pm$0.6 & 82.6$\pm$0.6          & 86.9$\pm$0.6          & 85.9$\pm$0.6          & 87.7$\pm$0.6          & 87.3$\pm$0.5          & \textbf{89.6$\pm$0.6} \\
\hdashline
Traffic Signs & \textbf{66.6$\pm$0.9} & 60.1$\pm$0.9          & 38.3$\pm$0.9 & 38.8$\pm$1.2 & 57.3$\pm$0.8 & 62.5$\pm$0.8          & \textbf{66.6$\pm$0.8} & 58.9$\pm$0.9          & 62.7$\pm$0.8          & 62.0$\pm$0.8          & \textbf{67.0$\pm$0.7} \\
MSCOCO        & 41.3$\pm$1.0          & \textbf{42.0$\pm$1.0} & 14.2$\pm$0.7 & 19.1$\pm$0.8 & 32.9$\pm$1.0 & 40.8$\pm$1.0          & \textbf{42.1$\pm$1.0} & 41.6$\pm$1.1          & 40.1$\pm$1.0          & 41.6$\pm$1.0          & \textbf{43.4$\pm$1.0} \\
MNIST         &92.1$\pm$0.4           & 88.6$\pm$0.5          & 65.9$\pm$0.8 & 82.5$\pm$0.6 & 86.8$\pm$0.5 & 89.8$\pm$0.5          & 91.3$\pm$0.4          & 92.1$\pm$0.4          & \textbf{93.2$\pm$0.3} & 90.5$\pm$0.4          & \textbf{92.3$\pm$0.4} \\
CIFAR10       & \textbf{70.1$\pm$0.8} & 60.0$\pm$0.8          & 26.1$\pm$0.7 & 29.1$\pm$0.6 & 55.8$\pm$0.8 & 65.9$\pm$0.8          & \textbf{69.7$\pm$0.7} & \textbf{69.6$\pm$0.8} & 66.9$\pm$0.8          & \textbf{70.3$\pm$0.8} & \textbf{69.3$\pm$0.8} \\
CIFAR100      & 55.6$\pm$1.0          & 48.1$\pm$1.0          & 16.7$\pm$0.8 & 16.7$\pm$0.7 & 37.9$\pm$1.0 & 52.9$\pm$1.0          & 55.0$\pm$1.0          & 54.2$\pm$1.1          & 53.0$\pm$1.1          & \textbf{59.5$\pm$1.0} & 54.6$\pm$1.1 \\
\midrule
Average Rank  & 3.92                  & 5.58                  & 10.69        & 10.31        & 7.96         & 7.54                  & 3.77                  & 4.04                  & 5.38                  & 4.42                  & 2.38 \\
\bottomrule
\end{tabular}
\end{adjustbox}
\end{small}
\end{center}

\caption[Few-shot classification results on \textsc{Meta-Dataset}]{Few-shot classification results on \textsc{Meta-Dataset} using the Prototypical Networks algorithm. Datasets below the dashed line are entirely held out. Meta-training performed on datasets above the dashed line. All figures are percentages and the $\pm$ sign indicates the 95\% confidence interval over tasks. Bold indicates the highest scores. Vertical lines in the TBN column indicate that this method is transductive.}
\label{tab:protonets_meta_dataset}
\begin{center}
\begin{small}
\begin{adjustbox}{max width=\textwidth}
\begin{tabular}{l | c | c c c c c c c c c c}
\toprule
Dataset       & TBN                   & CBN          & BRN                   & LN                    & IN           & RN                    & MetaBN                & TaskNorm-r   & TaskNorm-L            & TaskNorm-I \\
\midrule
ILSVRC        & \textbf{44.7$\pm$1.0} & 43.6$\pm$1.0 & 43.0$\pm$1.0          & 33.9$\pm$0.9          & 32.5$\pm$0.9 & \textbf{45.1$\pm$1.0} & \textbf{44.2$\pm$1.0} & 42.7$\pm$1.0 & \textbf{45.1$\pm$1.1} & \textbf{44.9$\pm$1.0} \\ 
Omniglot      & \textbf{90.7$\pm$0.6} & 77.5$\pm$1.1 & 89.1$\pm$0.7          & \textbf{90.8$\pm$0.6} & 83.4$\pm$0.8 & \textbf{90.8$\pm$0.6} & \textbf{90.4$\pm$0.6} & 88.6$\pm$0.7 & \textbf{90.2$\pm$0.6} & \textbf{90.6$\pm$0.6} \\
Aircraft      & 83.3$\pm$0.6          & 77.0$\pm$0.7 & \textbf{84.4$\pm$0.5} & 73.9$\pm$0.7          & 75.0$\pm$0.6 & 80.9$\pm$0.6          & 82.3$\pm$0.6          & 79.6$\pm$0.6 & 81.2$\pm$0.6          & \textbf{84.7$\pm$0.5} \\
Birds         & 69.6$\pm$0.9          & 67.5$\pm$0.9 & 69.0$\pm$0.9          & 54.1$\pm$1.0          & 50.2$\pm$1.0 & 68.6$\pm$0.9          & 68.6$\pm$0.8          & 64.2$\pm$0.9 & 68.8$\pm$0.9          & \textbf{71.0$\pm$0.9} \\
Textures      & 61.2$\pm$0.7          & 57.7$\pm$0.7 & 58.0$\pm$0.7          & 55.8$\pm$0.7          & 45.3$\pm$0.7 & 64.1$\pm$0.7          & 60.5$\pm$0.7          & 60.8$\pm$0.7 & 63.4$\pm$0.8          & \textbf{65.9$\pm$0.7} \\
Quick Draw    & 75.0$\pm$0.8          & 62.1$\pm$1.0 & 74.3$\pm$0.8          & 72.5$\pm$0.8          & 70.8$\pm$0.8 & 75.4$\pm$0.7          & 74.2$\pm$0.7          & 73.2$\pm$0.8 & 75.4$\pm$0.7          & \textbf{77.5$\pm$0.7} \\
Fungi         & 46.4$\pm$1.0          & 43.6$\pm$1.0 & 46.5$\pm$1.0          & 33.2$\pm$1.1          & 29.8$\pm$1.0 & 46.7$\pm$1.0          & 46.5$\pm$1.0          & 42.3$\pm$1.1 & 46.5$\pm$1.0          & \textbf{49.6$\pm$1.1} \\
VGG Flower    & 83.1$\pm$0.6          & 82.3$\pm$0.6 & 84.5$\pm$0.6          & 78.3$\pm$0.8          & 69.4$\pm$0.8 & 84.4$\pm$0.7          & \textbf{86.0$\pm$0.6} & 81.1$\pm$0.7 & 82.9$\pm$0.7          & 83.2$\pm$0.6 \\
\hdashline
Traffic Signs & 64.0$\pm$0.8          & 59.5$\pm$0.8 & 65.7$\pm$0.8          & \textbf{69.1$\pm$0.7} & 60.7$\pm$0.8 & 66.0$\pm$0.8          & 63.2$\pm$0.8          & 64.9$\pm$0.8 & 67.0$\pm$0.7          & 65.8$\pm$0.7 \\
MSCOCO        & 38.2$\pm$1.0          & 36.6$\pm$1.0 & \textbf{38.4$\pm$1.0} & 30.1$\pm$0.9          & 27.7$\pm$0.9 & 37.3$\pm$1.0          & \textbf{38.6$\pm$1.1} & 35.4$\pm$1.0 & \textbf{39.2$\pm$1.0} & \textbf{38.5$\pm$1.0} \\
MNIST         & 93.4$\pm$0.4          & 86.5$\pm$0.6 & 91.9$\pm$0.4          & \textbf{94.0$\pm$0.4} & 87.4$\pm$0.5 & \textbf{93.9$\pm$0.4} & \textbf{93.9$\pm$0.4} & 92.5$\pm$0.4 & 91.9$\pm$0.4          & 93.3$\pm$0.4 \\
CIFAR10       & 64.7$\pm$0.8          & 57.3$\pm$0.8 & 60.1$\pm$0.8          & 51.5$\pm$0.8          & 50.5$\pm$0.8 & 62.3$\pm$0.8          & 63.0$\pm$0.8          & 61.4$\pm$0.8 & \textbf{66.9$\pm$0.8} & \textbf{67.6$\pm$0.8} \\
CIFAR100      & 48.0$\pm$1.1          & 43.1$\pm$1.0 & 43.9$\pm$1.0          & 34.0$\pm$0.9          & 32.1$\pm$1.0 & 47.2$\pm$1.1          & 47.0$\pm$1.0          & 45.2$\pm$1.0 & \textbf{51.3$\pm$1.1} & \textbf{50.0$\pm$1.0} \\
\midrule
Average Rank  & 4.04                  &  8.19        &  5.31                 &  7.46                 &  9.58        & 3.65                  &  3.96                 & 6.73         &  3.58                 & 2.50 \\
\bottomrule
\end{tabular}
\end{adjustbox}
\end{small}
\end{center}

\end{table*}
Next, we evaluate NLs on a demanding few-shot classification challenge called Meta-Dataset, composed of thirteen (eight train, five test) image classification datasets \citep{triantafillou2019meta}.
Experiments are carried out with \cnaps{}, which achieves state-of-the-art performance on Meta-Dataset \citep{requeima2019cnaps} and ProtoNets.
The challenge constructs few-shot learning tasks by drawing from the following distribution. 
First, one of the datasets is sampled uniformly; second, the ``way'' and ``shot'' are sampled randomly according to a fixed procedure; third, the classes and context / target instances are sampled. 
As a result, the context size $D^\tau$ will vary in the range between 5 and 500 for each task. 
In the meta-test phase, the identity of the original dataset is not revealed and tasks must be treated independently (i.e.~no information can be transferred between them). 
The meta-training set comprises a disjoint and dissimilar set of classes from those used for meta-test. 
Details provided in \cref{app:experiment_details} and \citet{triantafillou2019meta}.

\paragraph{Accuracy}
The classification accuracy results for \cnaps{} and ProtoNets on Meta-Dataset are shown in \cref{tab:cnaps_meta_dataset} and \cref{tab:protonets_meta_dataset}, respectively.
In the case of ProtoNets, all the the NLs specifically designed for meta-learning scenarios outperform TBN in terms of classification accuracy based on their average rank over all the datasets. For \cnaps{}, both RN and \tasknorm{}-I meet or exceed the rank of TBN.
This may be as $|D^\tau|$
\begin{inlinelist}
    \item is quite large in Meta-Dataset, and
    \item may be imbalanced w.r.t. classes, making prediction harder with transductive NLs.
\end{inlinelist}
\tasknorm{}-I comes out as the clear winner ranking first in 11 and 10 of the 13 datasets using \cnaps{} and ProtoNets, respectively. 
This supports the hypothesis that augmenting the BN moments with a second, instance based set of moments and learning the blending factor $\alpha$ as a function of context set size is superior to fixing $\alpha$ to a constant value (as is the case with RN). 
With both algorithms, the instance based NLs fall short of the meta-learning specific ones. However, in the case of \cnaps{}, they outperform the running average based methods (CBN, BRN), which perform poorly. In the case of ProtoNets, BRN outperforms the instance based methods, and IN fairs the worst of all. In general, ProtoNets is less sensitive to the NL used when compared to \cnaps{}.
The \textsc{Baseline} column in \cref{tab:cnaps_meta_dataset} is taken from \citet{requeima2019cnaps}, where the method reported state-of-the-art results on Meta-Dataset.
The \textsc{Baseline} algorithm uses the running moments learned during pre-training of its feature extractor for normalization.
Using meta-learning specific NLs (in particular \tasknorm{}) achieves significantly improved accuracy compared to \textsc{Baseline}.

As an ablation, we have also added an additional variant of \tasknorm{} that blends the batch moments from the context set with the running moments accumulated during meta-training that we call \tasknorm{}-r. \tasknorm{}-r makes use of the global running moments to augment the local context statistic and it did not perform as well as the \tasknorm{} variants that employed local moments (i.e. \tasknorm{}-I and \tasknorm{}-L).

\paragraph{Training Speed}
\cref{fig:cnaps_training_curves} plots training loss versus training iteration for the models in \cref{tab:cnaps_meta_dataset} that use the \cnaps{} algorithm. 
The fastest training convergence is achieved by \tasknorm{}-I. 
The instance based methods (IN, LN) are the slowest to converge. 
Note that \tasknorm{} converges within 60k iterations while \textsc{Baseline} takes 110k iterations and IN takes 200k.
\cref{fig:proto_nets_meta_dataset_training_curves} shows the training curves for the ProtoNets algorithm. The convergence speed trends are very similar to \cnaps{}, with \tasknorm{}-I the fastest.

Our results demonstrate that \tasknorm{} is the best approach for normalizing tasks on the large scale Meta-Dataset benchmark in terms of classification accuracy and training efficiency.
Here, we see high sensitivity of performance across NLs. 
Interestingly, in this setting \tasknorm{}-I outperformed TBN in classification accuracy, as did both RN and \metanorm{}. 
This refutes the hypothesis that TBN will always outperform other methods due to its transductive property, and implies that designing NL methods specifically for meta-learning has significant value.
In general, the meta-learning specific methods outperformed more general NLs, supporting our third hypothesis.
We suspect the reason that \tasknorm{} outperforms other methods is due to its ability to adaptively leverage information from both $D^\tau$ and $\vx^{\tau\ast}$ when computing moments, based on the size of $D^\tau$.

\section{Conclusions}
\label{sec:conclusions}

We have identified and specified several issues and challenges with NLs for the meta-learning setting.
We have introduced a novel variant of batch normalization -- that we call \tasknorm{} -- which is geared towards the meta-learning setting. 
Our experiments demonstrate that \tasknorm{} achieves performance gains in terms of both classification accuracy and training speed, sometimes exceeding transductive batch normalization.
We recommend that future work in the few-shot / meta-learning community adopt \tasknorm{}, and if not, declare the form of normalization used and implications thereof, especially where transductive methods are applied.

\section*{Acknowledgments}
The authors would like to thank Elre Oldewage, Will Tebbutt, and the reviewers for their insightful comments and feedback. Richard E. Turner is supported by Google, Amazon, ARM, Improbable and EPSRC grants EP/M0269571 and EP/L000776/1.

%\clearpage
%\newpage
\bibliography{bibliography}
\bibliographystyle{icml2020}

\newpage
\appendix

\clearpage
\newpage

\appendix
\renewcommand\thefigure{\thesection.\arabic{figure}} 
\renewcommand\thetable{\thesection.\arabic{table}} 
\renewcommand\theequation{\thesection.\arabic{equation}} 
\renewcommand\thealgorithm{\thesection.\arabic{algorithm}}

\setcounter{figure}{0}
\setcounter{table}{0}
\setcounter{equation}{0}

\section{Additional Normalization Layers}
\label{app:more_nls}

Here we discuss various additional NLs that are relevant to meta-learning.

\subsection{Batch Renormalization (BRN)}
\label{app:batch_renorm}

Batch renormalization \citep[BRN;][]{ioffe2017batch} is intended to mitigate the issue of non-identically distributed and/or small batches while retaining the training efficiency and stability of CBN. 
In BRN, the CBN algorithm is augmented with an affine transform with batch-derived parameters which correct for the batch statistics being different from the overall population. The normalized activations of a BRN layer are computed as follows:
\begin{equation*}
\label{eq:brn_normalize}
    \va'_n = \vgamma \left( r \left( \frac{\va_n - \vmu_{BN}} { \vsigma_{BN} + \epsilon} \right) + d \right) + \vbeta, \quad 
\end{equation*}
where
\begin{align*}
r =& \texttt{stop\_grad} \left( \texttt{clip}_{[1/r_{max}, r_{max}]} \left(\frac{\vsigma_{BN}}{\vsigma_r} \right) \right), \\
d =& \texttt{stop\_grad} \left( \texttt{clip}_{[-d_{max}, d_{max}]} \left( \frac{\vmu_{BN} - \vmu_r}{\vsigma_r} \right) \right). \\
\end{align*}

Here $\texttt{stop\_grad}(\cdot)$ denotes a gradient blocking operation, and $\texttt{clip}_{\left[a, b \right]}$ denotes an operation returning a value in the range $\left[a, b\right]$.
Like CBN, BRN is not well suited to the meta-learning scenario as it does not map directly to the hierarchical form of meta-learning models.
In \cref{sec:experiments}, we show that using BRN can improve predictive performance compared to CBN, but still performs significantly worse than competitive approaches.
\cref{tab:maml_results} shows that batch renormalization performs poorly when using \textsc{MAML}.

\subsection{Group Normalization (GN)}
\label{app:group_norm}
A key insight of \citet{wu2018group} is that CBN performance suffers with small batch sizes.
The goal of Group Normalization \citep[GN;][]{wu2018group} is thus to address the problem of normalization of small batch sizes, which, among other matters, is crucial for training large models in a data-parallel fashion.
This is achieved by dividing the image channels into a number of groups $G$ and subsequently computing the moments for each group. 
GN is equivalent to LN when there is only a single group ($G = 1$) and equivalent to IN when the number of groups is equal to the number of channels in the layer ($G = C$).
\subsection{Other NLs}
\label{app:other_nls}
There exist additional NLs including Weight Normalization \citep{salimans2016weight}, Cosine Normalization \citep{luo2018cosine}, Filter Response Normalization \citep{singh2019filter}, among many others.

Weight normalization reparameterizes weight vectors in a neural network to improve the conditioning for optimization. 
Weight normalization is non-transductive, but we don't consider this approach further in this work as we focus on NLs that modify activations as opposed to weights. 

Filter Response Normalization (FRN) is another non-transductive NL that performs well for all batch sizes. 
However we did not include it in our evaluation as FRN also encompasses the activation function as an essential part of normalization making it difficult to be a drop in replacement for CBN in pre-trained networks as is the case for some of our experiments. 

Cosine normalization replaces the dot-product calculation in neural networks with cosine similarity for improved performance. 
We did not consider this method further in our work as it is not a simple drop-in replacement for CBN in pre-existing networks such as the ResNet-18 we use in our experiments.

\section{Experimental Details}
\label{app:experiment_details}

In this section, we provide the experimental details required to reproduce our experiments. 
The experiments using MAML \citep{finn2017model} were implemented in TensorFlow \citep{tensorflow2015-whitepaper}, the Prototypical Networks experiments were implemented in Pytorch \citep{paszke2017automatic}, and the experiments using \cnaps{} \citep{requeima2019cnaps} were implemented using a combination of TensorFlow \citep{tensorflow2015-whitepaper} and Pytorch.
All experiments were executed on NVIDIA Tesla P100-16GB GPUs.

\subsection{MAML Experiments}
\label{app:maml_experiments}

We evaluate MAML using a range of normalization layers on:
\begin{enumerate}
    \item Omniglot \citep{lake2011one}: a few-shot learning dataset consisting of 1623 handwritten characters (each with 20 instances) derived from 50 alphabets.
    \item miniImageNet \citep{vinyals2016matching}: a dataset of 60,000 color images that is sub-divided into 100 classes, each with 600 instances.
\end{enumerate}
For all the MAML experiments, we used the codebase provided by the MAML authors \citep{Finn2017code} with only small modifications to enable additional normalization techniques. 
Note that we used the first-order approximation version of MAML for all experiments.
MAML was invoked with the command lines as specified in the \texttt{main.py} file in the MAML codebase.
No hyper-parameter tuning was performed and we took the results from a single run.
All models were trained for 60,000 iterations and then tested. 
No early stopping was used. 
We did not select the model based on validation accuracy or other criteria. %
The MAML code employs ten gradient steps at test time and computes classification accuracy after each step. 
We report the maximum accuracy across those ten steps. 
To generate the plot in \cref{fig:maml_accuracy_vs_shot}, we use the same command line as Omniglot-5-1, but vary the update batch size from one to ten.
\subsection{\cnaps{} Experiments}
\label{app:cnaps_experiments}

We evaluate \cnaps{} using a range of normalization layers on a demanding few-shot classification challenge called Meta-Dataset \citep{triantafillou2019meta}. 
Meta-Dataset is composed of ten (eight train, two test) image classification datasets. 
We augment Meta-Dataset with three additional held-out datasets: MNIST \citep{lecun2010mnist}, CIFAR10 \citep{krizhevsky2009learning}, and CIFAR100 \citep{krizhevsky2009learning}. 
The challenge constructs few-shot learning tasks by drawing from the following distribution. 
First, one of the datasets is sampled uniformly; second, the ``way'' and ``shot'' are sampled randomly according to a fixed procedure; third, the classes and context / target instances are sampled. 
Where a hierarchical structure exists in the data (ILSVRC or \textsc{Omniglot}), task-sampling respects the hierarchy. 
In the meta-test phase, the identity of the original dataset is not revealed and the tasks must be treated independently (i.e.~no information can be transferred between them). 
Notably, the meta-training set comprises a disjoint and dissimilar set of classes from those used for meta-test. 
Full details are available in \citet{triantafillou2019meta}.

For all the \cnaps{} experiments, we use the code provided by the the \cnaps{} authors \citep{requeima2019code} with only small modifications to enable additional normalization techniques. 
We follow an identical dataset configuration and training process as prescribed in \citet{requeima2019code}. 
To generate results in \cref{tab:cnaps_meta_dataset}, we used the following \cnaps{} options: FiLM feature adaptation, a learning rate of 0.001, and TBN, CBN, BRN, and RN used 70,000 training iterations, IN used 200,000 iterations, LN used 110,000 iterations, and \tasknorm{} used 60,000 iterations. 
The \cnaps{} code generates two models: fully trained and best validation. We report the better of the two. 
We performed no hyper-parameter tuning and report the test results from the first run. 
Note that CBN, TBN, and RN share the same trained model. 
They differ only in how meta-testing is done.
\subsection{Prototypical Networks Experiments}
We evaluate the Prototypical Networks \citep{snell2017prototypical} algorithm with a range of NLs using the same Omniglot, miniImageNet, and Meta-Dataset benchmarks.

For Omniglot, we used the codebase created by the Prototypical Networks authors \citep{snell2017code}. For miniImageNet, we used the a different codebase (\citep{chen2017code}) as the first codebase did not support miniImageNet.
Only small modifications were made to the two codebases to enable additional NLs.
For Omniglot and miniImageNet, we set hyper-parameters as prescribed in \citep{snell2017prototypical}. Early stopping was employed and the model that produced the best validation was used for testing.

For Meta-Dataset, we use the code provided by the the \cnaps{} authors \citep{requeima2019code} with only small modifications to enable additional normalization techniques and a new classifier adaptation layer to generate the linear classifier weights per equation (8) in \citep{snell2017prototypical}.
We follow an identical dataset configuration and training process as prescribed in \citet{requeima2019code}. 
To generate results in \cref{tab:protonets_meta_dataset}, we used the following \cnaps{} options: no feature adaptation, a learning rate of 0.001, 60,000 training iterations for all NLs, and the pretrained feature extractor weights were not frozen and allowed to update during meta-training.

\section{Additional Classification Results}
\label{app:additional_classification_results}
\begin{figure*}
\centering
    \begin{subfigure}{0.48\textwidth}
    \centering
    \includegraphics[width=\columnwidth]{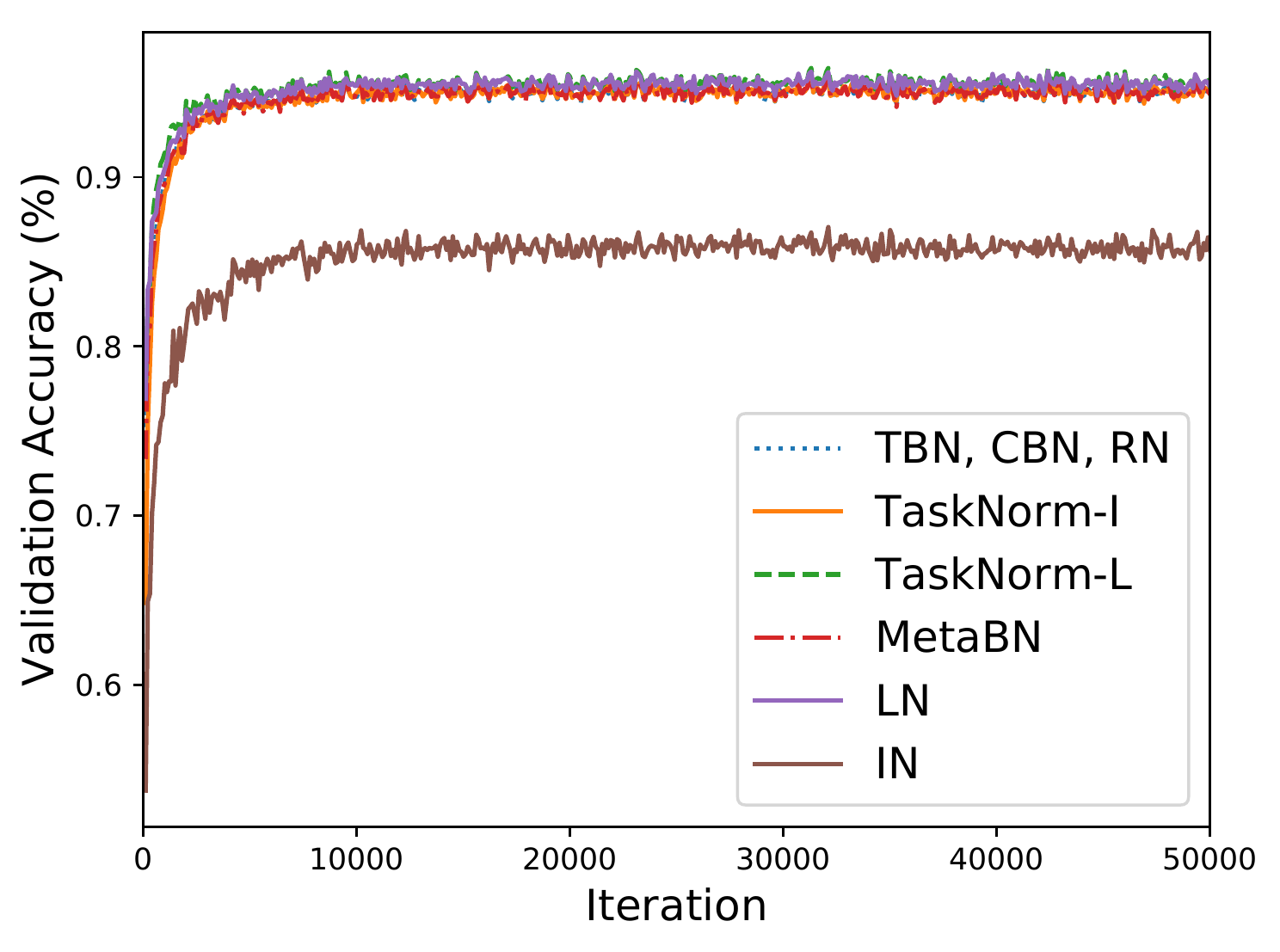}
    \caption{}
    \label{fig:proto_nets_o-20_1_training_curves}
    \end{subfigure}%
    \hfill
    \begin{subfigure}{0.48\textwidth}
    \centering
    \includegraphics[width=\columnwidth]{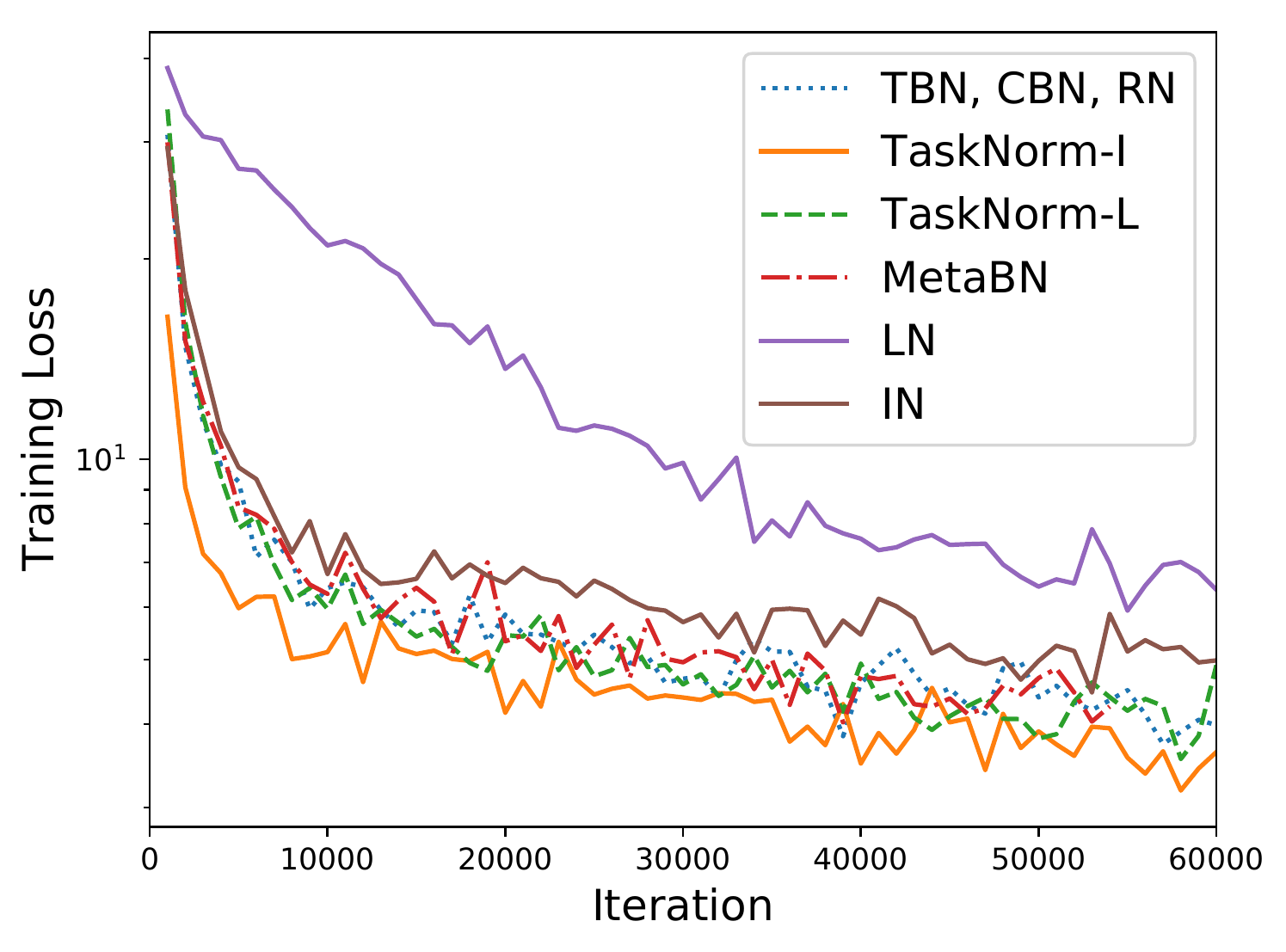}
    \caption{}
    \label{fig:proto_nets_meta_dataset_training_curves}
    \end{subfigure}%
    \caption{(a) Plot of validation accuracy versus training iteration using ProtoNets for Omniglot 20-way, 1-shot corresponding to the results in \cref{tab:protonets_results}. (b) Training Loss versus iteration corresponding to the results using the ProtoNets algorithm on \textsc{Meta-Dataset} in \cref{tab:protonets_meta_dataset}. Note that TBN, CBN, and RN all share the same meta-training step.}
\end{figure*}

\cref{tab:protonets_results} shows the classification accuracy results for the ProtoNets algorithm on the Omniglot and miniImageNet datasets. \cref{fig:proto_nets_o-20_1_training_curves} and \cref{fig:proto_nets_meta_dataset_training_curves} show the training curves for the ProtoNets algorithm on Omniglot and Meta-Dataset, respectively.

\begin{table*}[t]
\caption[Accuracy results for different few-shot settings on Omniglot and miniImageNet]{Accuracy results for different few-shot settings on Omniglot and miniImageNet using the Prototypical Networks algorithm. All figures are percentages and the $\pm$ sign indicates the 95\% confidence interval. Bold indicates the highest scores. The numbers after the configuration name indicate the way and shots, respectively. The vertical lines in the TBN column indicate that this method is transductive.}
\label{tab:protonets_results}
\begin{center}
\begin{small}
\begin{adjustbox}{max width=\textwidth}
\begin{tabular}{ l |c| c c c c c c c c}
\toprule
Configuration     & TBN                   & CBN                   & BRN                   & LN                    & IN           & RN                    & MetaBN                & TaskNorm-L            & TaskNorm-I \\
\midrule
Omniglot-5-1      & 98.4$\pm$0.2          & \textbf{98.5$\pm$0.2} & \textbf{98.5$\pm$0.2} & \textbf{98.7$\pm$0.2} & 93.7$\pm$0.4 & 98.0$\pm$0.2          & 98.4$\pm$0.2          & \textbf{98.6$\pm$0.2} & 98.4$\pm$0.2 \\ 
Omniglot-5-5      & \textbf{99.6$\pm$0.1} & \textbf{99.6$\pm$0.1} & \textbf{99.6$\pm$0.1} & \textbf{99.7$\pm$0.1} & 98.8$\pm$0.1 & \textbf{99.6$\pm$0.1} & \textbf{99.6$\pm$0.1} & \textbf{99.6$\pm$0.1} & \textbf{99.6$\pm$0.1} \\ 
Omniglot-20-1     & 94.5$\pm$0.2          & 94.5$\pm$0.2          & 94.6$\pm$0.2          & \textbf{94.9$\pm$0.2} & 83.5$\pm$0.3 & 94.1$\pm$0.2          & 94.5$\pm$0.2          & \textbf{95.0$\pm$0.2} & 93.4$\pm$0.2 \\ 
Omniglot-20-5     & \textbf{98.6$\pm$0.1} & \textbf{98.6$\pm$0.1} & \textbf{98.6$\pm$0.1} & \textbf{98.7$\pm$0.1} & 96.3$\pm$0.1 & \textbf{98.6$\pm$0.1} & \textbf{98.6$\pm$0.1} & \textbf{98.7$\pm$0.1} & \textbf{98.6$\pm$0.1} \\
miniImageNet-5-1  & 45.9$\pm$0.6          & \textbf{47.8$\pm$0.6} & 46.3$\pm$0.6          & \textbf{47.5$\pm$0.6} & 30.4$\pm$0.5 & 39.7$\pm$0.5          & 42.6$\pm$0.6          & \textbf{47.5$\pm$0.6} & 43.2$\pm$0.6 \\ 
miniImageNet-5-5  & 65.5$\pm$0.5          & \textbf{66.7$\pm$0.5} & 64.7$\pm$0.5          & \textbf{66.3$\pm$0.5} & 48.8$\pm$0.5 & 63.1$\pm$0.5          & 64.6$\pm$0.5          & 65.3$\pm$0.5          & 63.9$\pm$0.5 \\ 
\midrule
Average Rank      & 4.58                  & 3.25                  & 4.33                  & 2.75                  & 9.00         & 6.67                  & 5.25                  & 3.08                  & 6.08 \\
\bottomrule
\end{tabular}
\end{adjustbox}
\end{small}
\end{center}
\end{table*}

\section{Additional Transduction Tests}
\label{app:transduction_issues}

\begin{table*}[ht]
\caption{Few-shot classification results for TBN and \tasknorm{}-I on \textsc{Meta-Dataset} using the \cnaps{} algorithm. For each NL, the first column of results "All" reports accuracy when meta-testing is performed on the entire target set at once. The second column of results "Example" reports accuracy when meta-testing is performed one example at a time. The third column of results "Class" reports accuracy when meta-testing is performed one class at a time. All figures are percentages and the $\pm$ sign indicates the 95\% confidence interval over tasks. Meta-training is performed on datasets above the dashed line, while datasets below the dashed line are entirely held out.}
\label{tab:transduction_issues}
%\vskip 0.1in
\begin{center}
\begin{small}
\begin{adjustbox}{max width=\textwidth}
\begin{tabular}{l c c c | c c c}
\toprule
              & \multicolumn{3}{c}{TBN}                    & \multicolumn{3}{c}{\tasknorm{}-I}  \\
Dataset       & All          & Example      & Class        & All          & Example      & Class \\
\midrule
ILSVRC        & 50.2$\pm$1.0 & 9.5$\pm$0.3  & 11.8$\pm$0.4 & 50.4$\pm$1.1 & 50.4$\pm$1.1 & 50.4$\pm$1.1 \\ 
Omniglot      & 91.4$\pm$0.5 & 7.5$\pm$0.4  & 9.6$\pm$0.4  & 91.3$\pm$0.6 & 91.3$\pm$0.6 & 91.3$\pm$0.6 \\
Aircraft      & 81.6$\pm$0.6 & 11.8$\pm$0.4 & 14.4$\pm$0.4 & 83.8$\pm$0.6 & 83.8$\pm$0.6 & 83.8$\pm$0.6 \\
Birds         & 74.5$\pm$0.8 & 7.6$\pm$0.4  & 8.4$\pm$0.4  & 74.4$\pm$0.9 & 74.4$\pm$0.9 & 74.4$\pm$0.9\\
Textures      & 59.7$\pm$0.7 & 17.0$\pm$0.2 & 18.1$\pm$0.4 & 61.1$\pm$0.7 & 61.1$\pm$0.7 & 61.1$\pm$0.7 \\
Quick Draw    & 70.8$\pm$0.8 & 5.6$\pm$0.4  & 8.8$\pm$0.4  & 74.7$\pm$0.7 & 74.7$\pm$0.7 & 74.7$\pm$0.7 \\
Fungi         & 46.0$\pm$1.0 & 5.0$\pm$0.3  & 6.5$\pm$0.4  & 50.6$\pm$1.1 & 50.6$\pm$1.1 & 50.6$\pm$1.1 \\
VGG Flower    & 86.6$\pm$0.5 & 11.2$\pm$0.4 & 12.6$\pm$0.4 & 87.8$\pm$0.5 & 87.8$\pm$0.5 & 87.8$\pm$0.5 \\
\hdashline
Traffic Signs & 66.6$\pm$0.9 & 6.0$\pm$0.3  & 8.1$\pm$0.4  & 64.8$\pm$0.8 & 64.8$\pm$0.8 & 64.8$\pm$0.8 \\
MSCOCO        & 41.3$\pm$1.0 & 6.1$\pm$0.3  & 7.9$\pm$0.4  & 42.2$\pm$1.0 & 42.2$\pm$1.0 & 42.2$\pm$1.0 \\
MNIST         & 92.1$\pm$0.4 & 14.4$\pm$0.3 & 19.3$\pm$0.4 & 91.3$\pm$0.4 & 91.3$\pm$0.4 & 91.3$\pm$0.4 \\
CIFAR10       & 70.1$\pm$0.8 & 14.4$\pm$0.3 & 16.4$\pm$0.4 & 70.0$\pm$0.8 & 70.0$\pm$0.8 & 70.0$\pm$0.8 \\
CIFAR100      & 55.6$\pm$1.0 & 5.6$\pm$0.3   & 7.7$\pm$0.4  & 54.6$\pm$1.0 & 54.6$\pm$1.0 & 54.6$\pm$1.0 \\
\bottomrule
\end{tabular}
\end{adjustbox}
\end{small}
\end{center}
\end{table*}

A non-transductive meta-learning system makes predictions for a single test set label conditioned only on a single input and the context set. 
A transductive meta-learning system also conditions on additional samples from the test set.

\cref{tab:transduction_issues} demonstrates failure modes for transductive learning.
In addition to reporting the classification accuracy results when the target set is evaluated all at once (first column of results for each NL), we report the classification accuracy when meta-testing is performed one target-set \textit{example} at a time (second column of results for each NL), and one target-set \textit{class} at a time (third column of results for each NL).
\cref{tab:transduction_issues} demonstrates that classification accuracy drops dramatically for TBN when testing is performed one example or one class at a time.

Importantly, in the case of \tasknorm{}-I (or any non-transductive NL -- i.e. all of NLs evaluated in this work apart from TBN), the evaluation results are identical whether they are meta-tested on the entire target set at once, one example at a time, or one class at a time.
This shows that transductive learning is sensitive to the distribution over the target set used during meta-training, demonstrating that transductive learning is less generally applicable than non-transductive learning. 
In particular, transductive learners may fail to make good predictions if target sets contains a different class balance than what was observed during meta-training, or if they are required to make predictions for one example at a time (e.g. in streaming applications).

\section{Ablation Study: Choosing the best parameterization for $\alpha$}
\label{app:ablation_study}

There are a number of possibilities for the parameterization of the \tasknorm{} blending parameter $\alpha$. We consider four different configurations for each NL:
\begin{enumerate}
    \item $\alpha$ is learned separately for each channel (i.e. channel specific) as an independent parameter.
    \item $\alpha$ is learned shared across all channels as an independent parameter.
    \item $\alpha$ is learned separately for each channel (i.e. channel specific) as a function of context set size (i.e. $\alpha = \textsc{sigmoid}(\textsc{scale} |D^{\tau}| + \textsc{offset})$).
    \item $\alpha$ is learned shared across all channels as a function of context set size (i.e. $\alpha = \textsc{sigmoid}(\textsc{scale} |D^{\tau}| + \textsc{offset})$).
\end{enumerate}
\paragraph{Accuracy}
\cref{tab:parameterization_maml} and \cref{tab:parameterization_cnaps} show classification accuracy for the various parameterizations for MAML and the \cnaps{} algorithms, respectively using the \tasknorm{}-I NL. 

\begin{table}[t]
\caption{Few-shot classification results for two $\alpha$ parameterizations on Omniglot and miniImageNet using the MAML algorithm. All figures are percentages and the $\pm$ sign indicates the 95\% confidence interval over tasks. Bold indicates the highest scores.}
\label{tab:parameterization_maml}
\vskip 0.1in
\begin{center}
\begin{small}
\begin{adjustbox}{max width=\textwidth}
\begin{tabular}{ l c c }
\toprule
                  & \multicolumn{2}{c}{Independent} \\
Configuration     & Channel Specific      & Shared \\
\midrule
Omniglot-5-1      & 90.7$\pm$1.0          &  \textbf{94.4$\pm$0.8} \\ 
Omniglot-5-5      & 98.3$\pm$0.2          &  \textbf{98.6$\pm$0.2} \\ 
Omniglot-20-1     & \textbf{90.6$\pm$0.5} &  \textbf{90.0$\pm$0.5} \\ 
Omniglot-20-5     & \textbf{96.4$\pm$0.2} &  \textbf{96.3$\pm$0.2} \\ 
miniImageNet-5-1  & \textbf{42.6$\pm$1.8} &  \textbf{42.4$\pm$1.7} \\ 
miniImageNet-5-5  & \textbf{58.8$\pm$0.9} &  \textbf{58.7$\pm$0.9} \\
\midrule
Average Rank      & 1.67          & 1.33 \\
\bottomrule
\end{tabular}
\end{adjustbox}
\end{small}
\end{center}
\end{table}
\begin{table*}[ht]
\caption{Few-shot classification results for various $\alpha$ parameterizations on \textsc{Meta-Dataset} using the \cnaps{} algorithm. All figures are percentages and the $\pm$ sign indicates the 95\% confidence interval over tasks. Bold indicates the highest scores. Meta-training performed on datasets above the dashed line, while datasets below the dashed line are entirely held out.}
\label{tab:parameterization_cnaps}
%\vskip 0.1in
\begin{center}
\begin{small}
\begin{adjustbox}{max width=\textwidth}
\begin{tabular}{l c c c c}
\toprule
              & \multicolumn{2}{c}{Independent}                & \multicolumn{2}{c}{Functional}  \\
Dataset       & Channel Specific       & Shared                & Channel Specific      & Shared    \\
\midrule
ILSVRC        & 45.3$\pm$1.0           & \textbf{49.6$\pm$1.1} & \textbf{49.8$\pm$1.1} & \textbf{50.6$\pm$1.1} \\ 
Omniglot      & \textbf{90.8$\pm$0.6}  & \textbf{90.9$\pm$0.6} & \textbf{90.1$\pm$0.6} & \textbf{90.7$\pm$0.6} \\
Aircraft      & 82.3$\pm$0.7           & \textbf{84.6$\pm$0.6} & \textbf{84.4$\pm$0.6} & \textbf{83.8$\pm$0.6} \\
Birds         & 70.1$\pm$0.9           & 73.2$\pm$0.9          & 73.1$\pm$0.9          & \textbf{74.6$\pm$0.8} \\
Textures      & 54.8$\pm$0.7           & 58.5$\pm$0.7          & 61.0$\pm$0.8          & \textbf{62.1$\pm$0.7} \\
Quick Draw    & 73.0$\pm$0.8           & \textbf{73.9$\pm$0.7} & \textbf{74.2$\pm$0.7} & \textbf{74.8$\pm$0.7} \\
Fungi         & 43.8$\pm$1.0           & \textbf{47.6$\pm$1.0} & \textbf{48.0$\pm$1.0} & \textbf{48.7$\pm$1.0} \\
VGG Flower    & 85.9$\pm$0.6           & 86.3$\pm$0.5          & 86.5$\pm$0.7          & \textbf{89.6$\pm$0.6} \\
\hdashline
Traffic Signs & 62.6$\pm$0.8           & 62.6$\pm$0.8          & 60.1$\pm$0.8          & \textbf{67.0$\pm$0.7} \\
MSCOCO        & 38.3$\pm$1.1           & 40.9$\pm$1.0          & 40.2$\pm$1.0          & \textbf{43.4$\pm$1.0} \\
MNIST         & \textbf{92.6$\pm$0.4}  & 91.7$\pm$0.4          & 91.1$\pm$0.4          & \textbf{92.3$\pm$0.4} \\
CIFAR10       & 65.7$\pm$0.9           & 67.7$\pm$0.8          & 67.3$\pm$0.9          & \textbf{69.3$\pm$0.8} \\
CIFAR100      & 48.1$\pm$1.2           & 52.1$\pm$1.1          & \textbf{53.3$\pm$1.0} & \textbf{54.6$\pm$1.1} \\
\midrule
Average Rank  & 3.5                    & 2.5                   & 2.5                   & 1.5 \\
\bottomrule
\end{tabular}
\end{adjustbox}
\end{small}
\end{center}
\end{table*}

When using the MAML algorithm, there are only two options to evaluate as the context size is fixed for each configuration of dataset, shot, and way and thus we need only evaluate the independent options (1 and 2 above). 
\cref{tab:parameterization_maml} indicates that the classification accuracy for the channel specific and shared parameterizations are nearly identical, but the shared parameterization is better in the Omniglot-5-1 benchmark and hence has the best ranking overall.

When using the \cnaps{} algorithm on the Meta-Dataset benchmark, the best parameterization option in terms of classification accuracy is $\alpha$ shared across channels as a function of context size. 
One justification for having $\alpha$ be a function of context size can be seen in \cref{fig:a_b_curves}. 
Here we plot the line $\textsc{SCALE} |D^{\tau}| + \textsc{OFFSET}$ on a linear scale for a representative set of NLs in the ResNet-18 used in the \cnaps{} algorithm. 
The algorithm has learned that the \textsc{SCALE} parameter is non-zero and the \textsc{OFFSET} is almost zero in all cases. 
If a constant $\alpha$ would lead to better accuracy, we would see the opposite (i.e the \textsc{SCALE} parameter would be at or near zero and the \textsc{OFFSET} parameter being some non-zero value). 
From \cref{tab:parameterization_cnaps} we can also see that accuracy is better when the parameterization is a shared $\alpha$ opposed to having a channel-specific $\alpha$.

\begin{figure*}
\centering
    \begin{subfigure}{0.48\textwidth}
    \centering
    \includegraphics[width=\columnwidth]{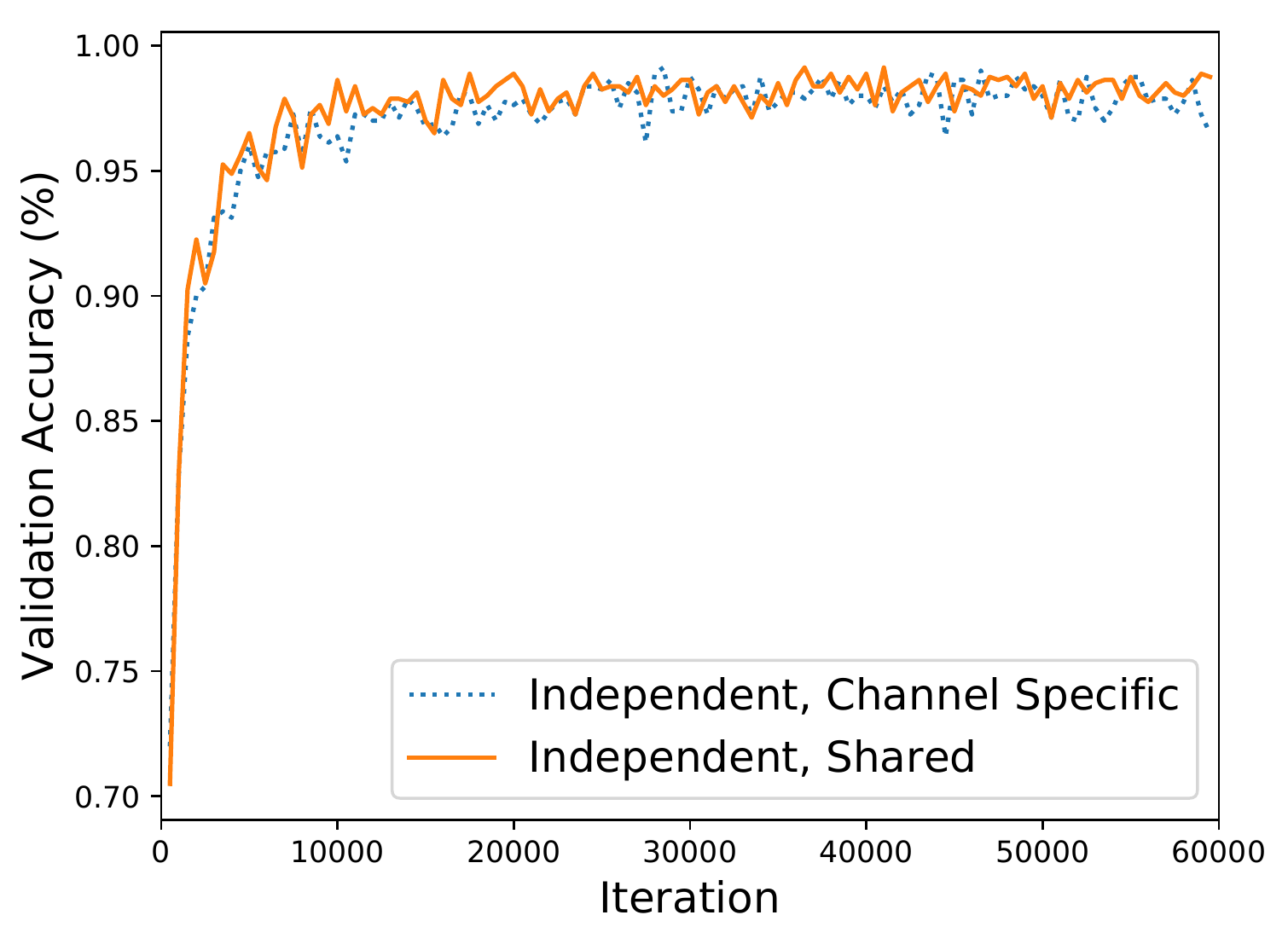}
    \caption{}
    \label{fig:maml_parameterization_training_curves}
    \end{subfigure}%
    \begin{subfigure}{0.48\textwidth}
    \centering
    \includegraphics[width=\columnwidth]{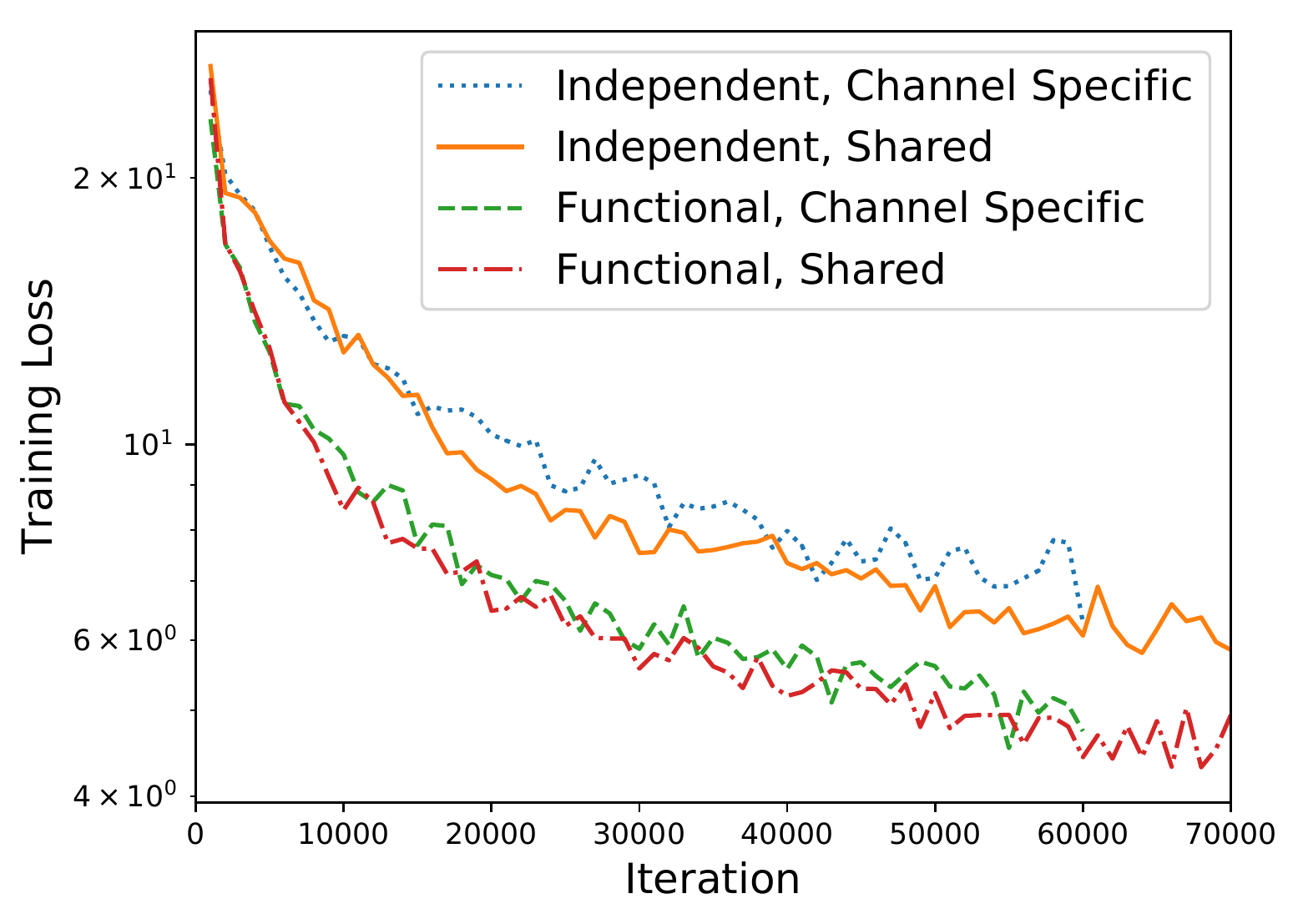}
    \caption{}
    \label{fig:cnaps_parameterization_training_curves}
    \end{subfigure}
    \caption{(a) Plots of validation accuracy versus training iteration corresponding to the parameterization experiments using the MAML algorithm in \cref{tab:parameterization_maml}. (b) Plot of training loss versus iteration corresponding to the parameterization experiments using the CNAPS algorithm in \cref{tab:parameterization_cnaps}.}
\end{figure*}

\paragraph{Training Speed}
\cref{fig:maml_parameterization_training_curves} and \cref{fig:cnaps_parameterization_training_curves} show the learning curves for the various parameterization options using the MAML and the \cnaps{} algorithms, respectively with a \tasknorm{}-I NL.

For the MAML algorithm the training efficiency of the shared and channel specific parameterizations are almost identical.
For the \cnaps{} algorithm, \cref{fig:cnaps_parameterization_training_curves} indicates the training efficiency of the independent parameterization is considerably worse than the functional one. 
The two functional representations for the CNAPs algorithm have almost identical training curves. 
Based on \cref{fig:maml_parameterization_training_curves} and \cref{fig:cnaps_parameterization_training_curves}, we conclude that the training speed of the functional parameterization is superior to that of the independent parameterization and that there is little or no difference in the training speeds between the functional, shared parameterization and the functional, channel specific parameterization.

In summary, the best parameterization for $\alpha$ when it is learned shared across channels as a function of context set size (option 4, above). 
We use this parameterization in all of the \cnaps{} experiments in the main paper. For the MAML experiments, the functional parameterization is meaningless given that all the test configurations have a fixed context size. 
In that case, we used the independent, shared across channels parameterization for $\alpha$ for the experiments in the main paper.

\end{document}